\documentclass[runningheads]{llncs}
\usepackage{graphicx}
\usepackage{cite}
\usepackage{algorithm}
\usepackage{algorithmic}
\usepackage{xcolor}
\usepackage{chngpage}
\usepackage{amsmath}
\usepackage{mathtools}
\usepackage{cleveref}
\usepackage{adjustbox}
\usepackage{booktabs}
\usepackage{adjustbox}
\usepackage{rotating}

\usepackage{color,soul}

\usepackage{caption}
\usepackage{subcaption}

\begin{document}
\long\def\/*#1*/{}

\title{Optimising Communication Overhead in Federated Learning Using NSGA-II}
\titlerunning{Optimising Communication Overhead in Federated Learning using NSGA-II}
% $\dagger$  $\ddagger$
\author{José Ángel Morell $\dagger$ \and Zakaria Abdelmoiz Dahi $\dagger$$\ddagger$\and Francisco Chicano $\dagger$\and Gabriel Luque $\dagger$\and Enrique Alba $\dagger$}
\authorrunning{J.A. Morell, Z.A. Dahi, F. Chicano, G. Luque and E. Alba}
\institute{$\dagger$ ITIS Software, University of Malaga, Spain.\\
$\ddagger$ Dep. Fundamental Computer Science and Its Applications, Fac. NTIC, \\
University of Constantine 2, Algeria. \\
\email{\{jamorell,chicano,gabriel,eat\}@lcc.uma.es}\\
\texttt{zakaria.dahi@\{uma.es, univ-constantine2.dz\}}}
\maketitle              % typeset the header of the contribution
\begin{abstract}
Federated learning is a training paradigm according to which a server-based model is cooperatively trained using local models running on edge devices and ensuring data privacy. These devices exchange information that induces a substantial communication's load, which jeopardises the functioning efficiency. The difficulty of reducing this overhead stands in achieving this without decreasing the model's efficiency (contradictory relation). To do so, many works investigated the compression of the pre/mid/post-trained models and the communication rounds, separately, although they jointly contribute to the communication overload. Our work aims at optimising communication overhead in federated learning by (I) modelling it as a multi-objective problem and (II) applying a multi-objective optimization algorithm (NSGA-II) to solve it. To the best of the author's knowledge, this is the first work that \texttt{(I)} explores the add-in that evolutionary computation could bring for solving such a problem, and \texttt{(II)} considers both the neuron and devices features together. We perform the experimentation by simulating a server/client architecture with 4 slaves. We investigate both convolutional and fully-connected neural networks with 12 and 3 layers, 887,530 and 33,400 weights, respectively. We conducted the validation on the \texttt{MNIST} dataset containing 70,000 images. The experiments have shown that our proposal could reduce the communication by 99\% and maintain an accuracy equal to the one obtained by the FedAvg Algorithm that uses 100\% of communications.    
\keywords{Federated Learning  \and Evolutionary Computation \and Multi-objective Optimisation}
\end{abstract}

\vspace{-2em}
%https://www.cisco.com/c/dam/m/en_us/service-provider/ciscoknowledgenetwork/files/622_11_15-16-Cisco_GCI_CKN_2015-2020_AMER_EMEAR_NOV2016.pdf
%https://blogs.cisco.com/sp/five-things-that-are-bigger-than-the-internet-findings-from-this-years-global-cloud-index
\section{Introduction \label{sec:intr}}

Today's advances in Artificial Intelligence (AI) allow training machine learning models by exploiting the daily-generated data that was previously considered useless~\cite{lecun2015deep}. Statista\footnote{\url{www.statista.com/statistics/1101442/iot-number-of-connected-devices-worldwide}} has stated that there are 23.8 billion interconnected computing devices that are active in the world and will produce 149 zettabytes of data by 2024. Cisco\footnote{\url{blogs.cisco.com/sp/five-things-that-are-bigger-than-the-internet-findings-from-this-years-global-cloud-index}} also estimated that at least 85 (10\%) of the 850 zettabytes created in 2021 will be useful, while only 7 zettabytes of it will be stored.  Indeed, most of this data cannot be stored/processed on the cloud despite the exponential increase of data demand and generation speed. In addition, data privacy prevents sharing it with third-parties (e.g. medical images). As promising solution to these issues, Federated Learning (FL) appeared. It is a learning paradigm that trains a shared model in a distributed manner while keeping private the data locally on edge devices. Federated Learning is being actively investigated and widely applied (e.g. medicine~\cite{sheller2020federated}). Its working mechanism induces a substantial communication overload that limits its applicability. It has been proven that this overhead is generated by several factors such as the number of devices participating in the learning process, the complexity of the model (e.g. number of layers, neurons, etc.), number of communication rounds, etc.~\cite{mayer2020scalable}. Previous works have already investigated some of these factors in isolation to decrease the communication excess~\cite{entry_quantisation,entry_sparsification,xu2020ternary}, although, one should note that the factors are jointly contributing to the communication overhead.

% Motivate the use of NSGA-II and contribution
Achieving high-quality results requires performing efficient network training using substantial information and  communication~\cite{mayer2020scalable}. Thus, the main difficulty when reducing the communication cost stands in maintaining the same efficacy, due to the conflictual relation between both. Using classical exhaustive tools turns out to be computationally costly and time-consuming, due to the complexity of the problem and its multi-criterion nature. For such problem's class, stochastic algorithms such as metaheuristics and, in particular, the Non-Dominated Sorting Genetic Algorithm II (NSGA-II) are a promising alternative that provides a good trade-off between the solving efficiency and time consumption~\cite{DPA02}. Bearing in mind the above-stated facts, our contributions stand in \texttt{(I)} modelling and formulating the Federated Learning Communication Overhead as a multi-objective Problem (FL-COP), \texttt{(II)} applying NSGA-II to solve it and \texttt{(III)} investigating within the same work the main parameters triggering communication overhead that the literature usually tackle separately. Our proposal has been assessed by simulating a server/client architecture of 4 devices, each one being tested with both convolutional and fully connected neural networks with 12 and 3 layers, 887,530 and 33,400 weights, respectively. The validation has been done on \texttt{MNIST} dataset containing 70,000 images of handwritten digits.  

% Reminder
The rest of the paper is structured as follows. In Section \ref{sec:fc}, we present basic concepts of FL and communication-overhead reduction strategies. Section \ref{sec:pa} introduces our proposal as well as our FL-COP formulation. Section \ref{sec:esa} presents the experimental results and analysis. Finally, Section \ref{sec:cp} concludes our work. 

%\vspace{-1.5em}

\section{Fundamental Concepts \label{sec:fc}}

%\vspace{-1em} %Dangerous! I edited a small thing above and text overlapped in the PDF.

This section presents the basic concepts of federated learning and the communication reduction strategies.

%\vspace{-1em}

\subsection{Federated Learning \label{sec:FL}}

\begin{figure}[!htb]
     \centering
     \includegraphics[scale=0.21]{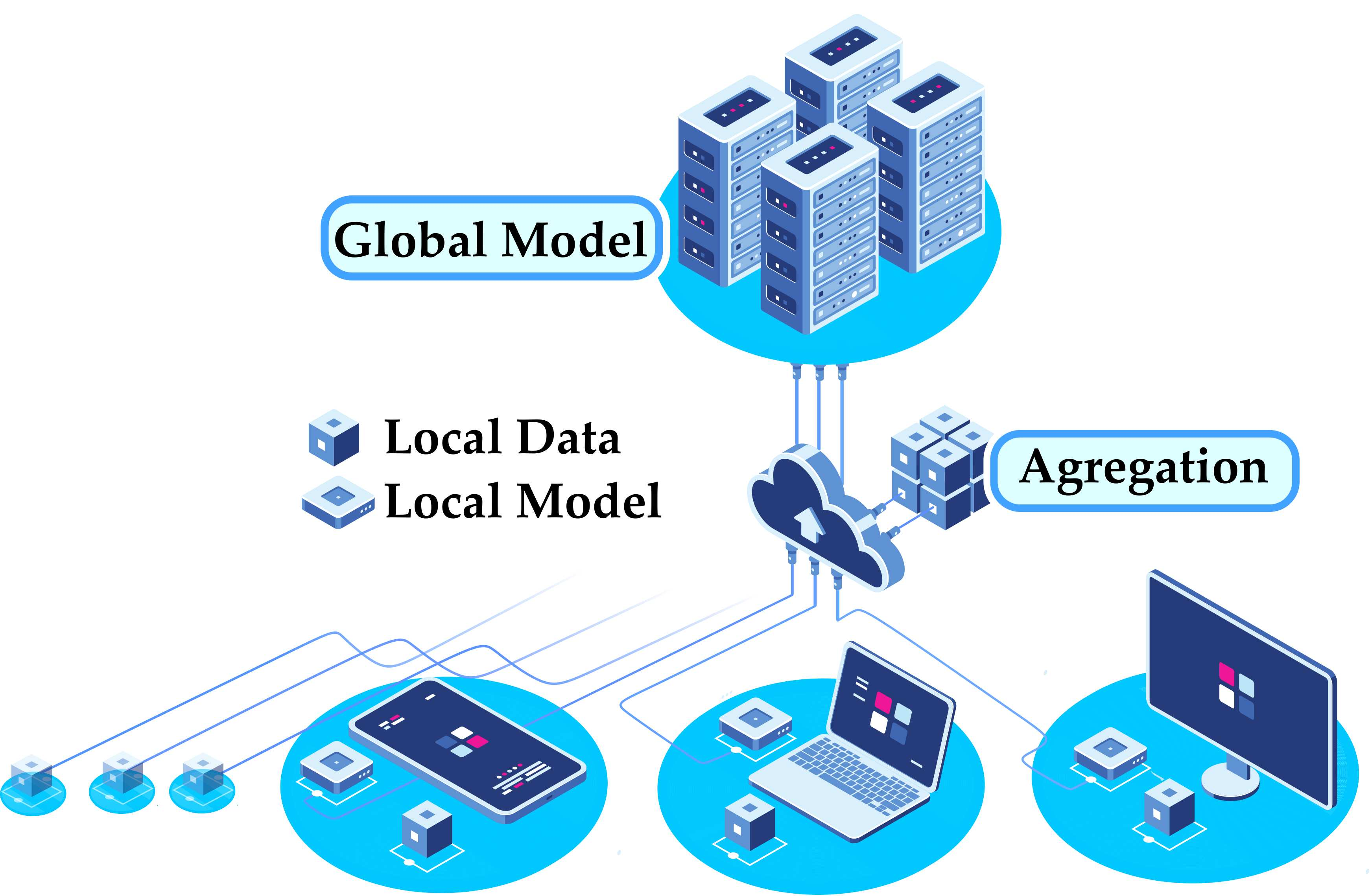}
     \caption{Federated learning architecture.}\label{fig:fl_arch}
\end{figure}

Originated from distributed deep learning~\cite{mayer2020scalable}, FL allows training a common model without compromising the users' data privacy. The latter are kept on local devices during the learning process. Instead of sharing the training data, the clients exchange their local models to help improve a global one (see Fig. \ref{fig:fl_arch}).

\vspace{-2em}

\vspace{-1.5em}

Algorithm (\ref{alg:fedavg}) sketches how the FederatedAveraging algorithm (\texttt{FedAVG})\cite{mcmahan2017communication}  proceeds using a cluster of \textit{N} clients, each with a learning rate of $\eta$. The variable \textit{S} designates the set containing all clients, while \textit{C} is a fraction representing a subset of selected clients from $S$, where $|S'| = (C \cdot N)$. The \texttt{FedAVG} acts in two synchronous steps, starting by generating a global model, say $w_0$, on the server. After that, it randomly chooses $m$ participating clients where $m$ is the maximum between ($C \cdot N$) and 1. Each of the selected clients trains a local model similar to the global one during several local iterations $e = 1, \dots, E$, where \textit{E} is the communication interval. Once done, all local models are sent to the server in order to update the global one, where $P_k$ is the weight of the $k^{th}$ client. The whole process is executed repeatedly during $T$ iterations. 

\vspace{-1em}

\begin{algorithm}[h!] 
 \begin{algorithmic}[1]
\STATE \texttt{Initialise}($w_0$);
\FOR{$t = 1, ..., T$}
\STATE $m \leftarrow \texttt{max}(C \cdot N, 1)$;
\STATE $S' \leftarrow \texttt{random\_Pick}(S, m)$;
\FORALL{clients $k \in S'$ in parallel}
\FOR {$e \in 1, ..., E$}
\STATE $w_e \leftarrow w_{e-1} - \eta \nabla F(w_{e-1})$;
\ENDFOR
\STATE $w_{t+1}^k \leftarrow w_e$;
\ENDFOR
\STATE $w_{t+1} \leftarrow \sum_{k=0}^{m} {P_{k} \cdot w_{t+1}^k}/m$;
\ENDFOR
\caption{The federated averaging algorithm.}
\label{alg:fedavg}
\end{algorithmic}
\end{algorithm}

\vspace{-3em}

\subsection{Communication Overhead in Distributed Deep Learning \label{sec:coddl}}

\begin{figure}
    \centering
    \includegraphics[scale = 0.26]{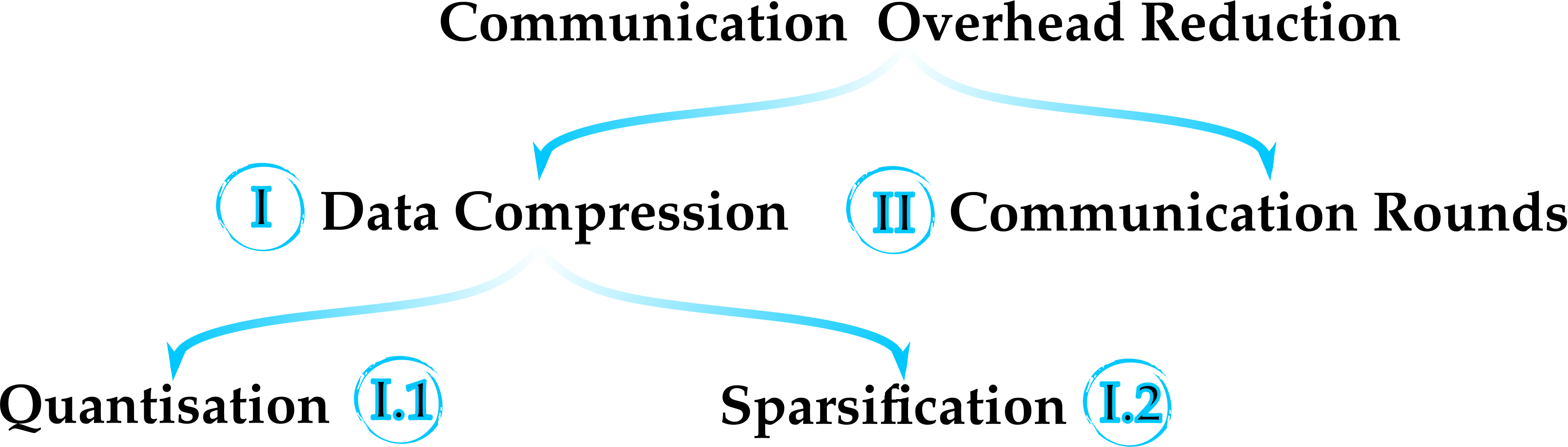}
    \caption{Communication overhead reduction: taxonomy of the techniques.}
    \label{fig:classification}
\end{figure}

The FL workflow, like any Distributed Deep Learning (DDL), induces a substantial load of communication, which decreases its efficiency and applicability \cite{tak2020federated}. When going through the DDL literature, two main approaches exist for communication overhead reduction: \texttt{(I)} \emph{data compression} and \texttt{(II)} \emph{decreasing communication rounds} (see Fig. \ref{fig:classification}).

\vspace{-1em} 
 
\vspace{-1em}

Considering the first approach, the literature identifies two effective ways for compressing data: \texttt{(I.1)} quantisation \cite{entry_quantisation} and \texttt{(I.2)} sparsification \cite{entry_sparsification}. The first consists of representing the data using a low-precision/small-sized data type (e.g. bool). In contrast, the second approach transmits only essential values of each communication (about 1\% of the overall values). Nonetheless, when using quantisation, the compression rate is low considering that the maximum compression ratio is limited to $(1/32)$ (32-bit-encoded data is frequently used in DDL). Also, when having fewer bits to carry the information, the models that use quantisation tend to have a slower convergence. Unlike quantisation, sparsification achieves a compression rate of $(1/100)$ without a significant modification of the model's convergence speed and final accuracy. Sparsification also comes backhanded since it introduces supplementary phases during the training process (e.g. sampling, de/compression, de/coding, etc.). This can affect the overall training efficiency, especially in battery-sensitive (e.g. smartphone) and low-performance (e.g. netbook) devices.

Moving now to the second communication-reduction technique, in vanilla FL~\cite{zhou2021communication} (i.e. standard FL), the communication happens at the end of each iteration ($E$ = 1). A typical FL training of deep neural network takes hundreds of thousands of iterations. Enlarging the communication intervals would allow reducing the communication overhead. Therefore, \texttt{FedAvg} algorithm and its variants allow clients to perform multiple iterations of local training before updating the global model~\cite{zhou2021communication}. It has been proven that reducing the communication rounds increases the convergence speed. The communication interval in \texttt{FedAvg} is controlled by the hyperparameter $E$, which influences the model's trade-off between the accuracy and the training efficiency. Generally, a smaller $E$ induces a better final accuracy, while a large $E$ value accelerates the model's convergence. Therefore, experts would be needed to fine-tune the communication interval $E$ to allow the model to reach the best possible efficacy. \\ 

Most of the literature studies the communication-reduction approaches separately. Although, we believe that they are all equally important and jointly impact the communication rate. Therefore, as far as the authors' knowledge, we are the first to investigate all these approaches together within the same work.  

\vspace{-1em}

\section{Proposed Approach \label{sec:pa}}

This section presents our FL-COP formulation and the used NSGA-II solver.

\vspace{-1em}

\subsection{The Proposed FL-COP Modelling and Formulation \label{sc:ff}}

\begin{figure}
    \centering
    \includegraphics[scale=0.36]{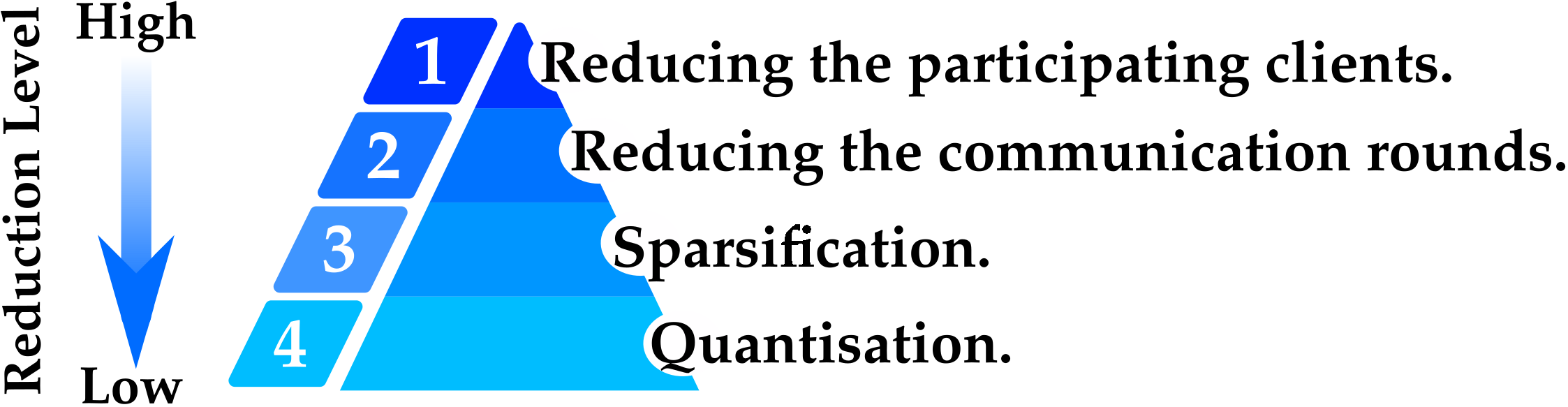}
    \caption{The FL-COP modelling levels}
    \label{fig:crl}
\end{figure}

Our formulation of the FL-COP is a bi-objective optimisation problem, where the two conflictual objectives consist of \texttt{(I)} minimising the communication overhead while \texttt{(II)} maximising the model's accuracy. It also assumed that each client in the architecture has a similar model (i.e. nodes, connections, layers, activation functions, etc.) as the one on the server. When mentioning the \emph{local} and \emph{global} models, we refer to the client's and server's models, respectively. Let us assume an architecture of one server connected to $N$ clients. The model being trained has $l$ layers $L_{i}$ having $n_{i}$ weights, where $i = {1,\ldots, l}$. The FL-COP modelling is thought as a 4-levels communication-reduction scheme, where each layer represents when a given communication-reduction approach is applied. At the highest level, we identify the number of clients that will participate in training the global model, while the three remaining lower levels reflect the three communication-reduction approaches explained in Section \ref{sec:coddl}: quantisation, sparsification and reducing the communication rounds (see Fig. \ref{fig:crl}).  

\vspace{-1.25em}

\vspace{-1.5em}

The overall amount of communications happening during the FL learning process is proportional to the number of clients $m \in [1, N]$ that participate in training the global model. So, a first part of the FL-COP modelling stands in finding the number $m$ of clients, selected randomly among the complete set $S$, and which will be the only ones sending their local models to the server. A second part of the FL-COP modelling consists in finding the number of training iterations $E \in [1, 1000]$ after which all the clients send their local models to the server. This variable determines the number of training steps that the clients perform before sending their local models (e.g. weights, gradients, etc.). It is important to note that for each client, the maximum number of training iterations allowed on overall is ($E \cdot T$), where $T$ is the maximum number of times the clients can send their local models to the server. The third part of the problem modelling consists in selecting, for each layer $L_{i}$ having $n_i$ weights, a percentage $\mu \in [0\%,50\%]$ of the weights that will not be sent to the server.  \\

Using the classical \texttt{FedAvg}, the weights are encoded with full precision (i.e. all their decimals) using 32 bits. Thus, the fourth, and final part of our FL-COP modelling consists in finding the optimal number of bits $b_i$ allocated to encode the weights of each layer $L_{i}$ in the model, where $i = 1, \dots, l$. We also assume that $\varpi_{i}$ and $\varrho_{i}$ represent, respectively, the maximum and minimum values of the weights in the $i^{th}$ layer. Having $b_{i}$ bits means that $2^{b_{i}}$ binary combinations can be created. We assign the all-ones and all-zeros combinations to encode the $\varpi_{i}$ and $\varrho_{i}$ values, respectively. The ($2^{b_{i}}$ - 2) remaining combinations will encode ($2^{b_{i}}$ - 2) values that are equally drawn from the interval [$\varpi_{i}$, $\varrho_{i}$]. Technically, the data that will be sent to the server will be the series of combinations that encodes each weight, as well as $\varpi_{i}$ and $\varrho_{i}$. The server will perform the reverse mechanism to retrieve the full-precision weights. Each client will send to the server $\sum_{i=1}^{l} (n_{i} \, . \, b_{i}) + 64$ bits instead of $\Theta = \sum_{i=1}^{l}{ (n_{i} \, . \, 32)}$ original bits.        

Our formulation of the FL-COP is described using Equations~\eqref{eq:1}-\eqref{eq:3}. The first objective function $f{_1}(\overrightarrow{X})$ defined by Equation~\eqref{eq:1} calculates the percentage of data reduction that the solution $\overrightarrow{X}$ achieves. Concretely, it is the sum of the percentage $\alpha$ and $\beta \in [0,1]$ of data sent and received, respectively, by all the clients together from and to the server. These percentages are expressed with regard to the original data that would have been sent or received when no communication reduction is applied $(T \cdot N \cdot \Theta)$. The second objective function $f_2(\overrightarrow{X})$ defined by Equation~\eqref{eq:2} evaluates the accuracy of the global model $w_{T}^{*}$ at communication $T$ (i.e. the last iteration) achieved via the solution $\overrightarrow{X}$. The server's model $w_{T}^{*} = \sum_{k=0}^{m} {{w_{T}^k}/{m}}$ is computed as the mean of the $m$ local models obtained after $T$ communications, while the accuracy is computed as the division of $\lambda$ by $\nu$, where $\lambda$ and $\nu$ are the number of correct and total predictions made using the model $w_{T}^{*}$, respectively. 

\begin{equation}
    \underset{\overrightarrow{X}=\{x_1, \dots, x_d\}}{\texttt{Min}} \, \, \, \, \, f_{1}(\overrightarrow{X}) = \frac{\alpha + \beta}{2}
    \label{eq:1}
\end{equation}

\begin{equation}
    \underset{\overrightarrow{X}=\{x_1, \dots, x_d\}}{\texttt{Max}} \, \, \, \, \, f_{2}(\overrightarrow{X}) = \frac{\lambda}{\nu}
    \label{eq:2}
\end{equation}

\texttt{Where:}
\vspace{-1em}
\begin{equation}
 \alpha = \frac{1}{E} \cdot \frac{m}{N}
\end{equation}

\begin{equation}
   \beta =  \frac{m}{N} \cdot \frac{1}{E} \cdot \sum_{i=1}^{l} \frac{b_{i}}{32} \cdot \frac{100 - \mu_{i}}{100} \cdot \frac{n_{i}}{\sum_{j=1}^{l} n_{j}}
    \label{eq:3}
\end{equation}

\texttt{Subject to:}  
\begin{center}
$m, E, \mu_i, b_i \in \mathcal{N}, 1 \leq m \leq N, 1 \leq E \leq 1000, 0 \leq \mu_i \leq 50, 1 \leq b_i \leq 32$ 
\end{center}

\begin{figure}
    \centering
    \includegraphics[scale = 0.28]{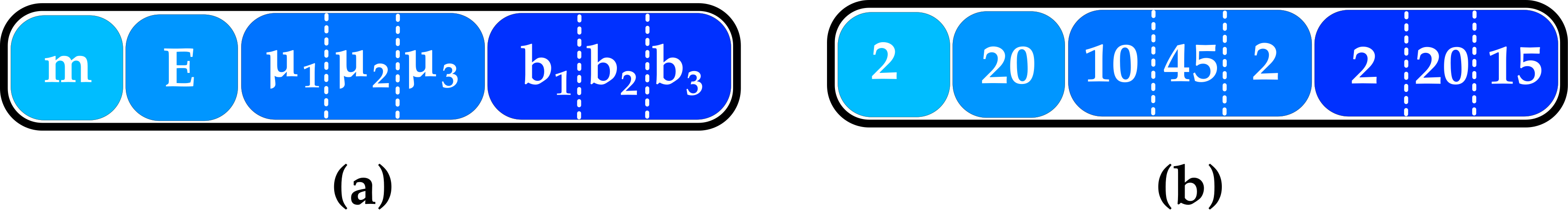}
    \vspace{-.5em}
    \caption{A 3-layers model: (a) abstract and (b) concrete FL-COP solutions.}
    \label{fig:ss}
\end{figure}

The Fig. \ref{fig:ss}(a) sketches a typical solution $\overrightarrow{X}$  of an FL-COP that trains a $l$ = 3 layers model. On the other hand, Fig. \ref{fig:ss}(b) represents a concrete solution $\overrightarrow{X}$ for the same configuration using 20 training iterations, 2 rounds of client-server communications. During each round, only 90\% of the weights of the $1^{st}$ layer, 55\% of the $2^{nd}$, and 98\% from the $3^{rd}$ are sent to the server. The weights sent from the $1^{st}$, $2^{nd}$ and $3^{rd}$ layers are encoded using 2, 20 and 15 bits, respectively. 

\vspace{-0.75em}

\vspace{-0.5em}

\subsection{The Communication-Overhead Reduction Routine}

To solve the FL-COP presented in Section \ref{sc:ff}, our approach consists in applying NSGA-II, a well-known evolutionary algorithm proposed by Deb et al.~\cite{DPA02} and initially designed to tackle multi-objective problems. NSGA-II main contributions are the non-dominated sorting and the diversity-preservation heuristics with a computational complexity of $\mathcal{O}$($MN^2$) and $\mathcal{O}$($MN$\texttt{log}$N$). Having a problem with $M$ objectives, NSGA-II starts by randomly initialising a population of $U$ individuals, let us say $\overrightarrow{X}$=\{$x_i$, $\dots$, $x_d$\}, where $i\in[1,d]$ and $d$ is the size of the problem to be solved. Once this is done, NSGA-II enters in a loop until some stopping criterion is fulfilled. In the loop, it applies binary tournament selection,  crossover and mutation to generate a population $Q$ of $U$ offspring. The union of both the parent and offspring populations, $R$ = $P \cup Q$, will be used as input of  a replacement operator in order to decide the solutions of the new population $P'$ that will survive to the next iteration  (see Algorithm \ref{alg:nsga-II}). 

\vspace{-1em}

\begin{algorithm}[H]
\caption{The non-dominated sorting genetic algorithm II. 
\label{alg:nsga-II}}
\begin{algorithmic}[1]
\STATE Set $M$ objective functions $O_i$/ $i \in \{1, \dots, M$\}.
\STATE Set $\overrightarrow{X}$ a typical solution/ $\overrightarrow{X}$=\{$x_1$, $\dots$, $x_d$\}, $d$ number of variables to optimise.   
\STATE Set $F= \{F_1, \dots, F_K$\}, $K$ the number of non-dominated fronts in the population. 
\STATE $P \leftarrow $ \texttt{Random\_Generation}($U$);\\
\WHILE{stopping criterion is not reached yet}
\STATE $A \leftarrow$ \texttt{Binary\_Tournament\_Selection}($P$, \texttt{Crowded\_Comparison});\\
\STATE $B \leftarrow$ \texttt{Crossover}($A$);
\STATE $Q \leftarrow$ \texttt{Mutation}($B$);\\
\STATE $F \leftarrow$ \texttt{Non\_Dominated\_Sorting}(P $\cup$ Q);
\STATE $P'$ $\leftarrow \, \emptyset$; 
\STATE $i \longleftarrow 1$;
\WHILE{($|P' \cup F{_i}| \leq U$ and $i \leq K$)}
\STATE $P' \leftarrow P' \cup F_i$;
\STATE $i \leftarrow i + 1 $;
\ENDWHILE
\STATE $F_i \leftarrow \texttt{Descending\_Sort\_Crowding\_Comparison}(F_i)$;
\STATE $P \leftarrow P' \cup F_i[1:(U - |P'|)]$;
\ENDWHILE
\end{algorithmic}
\end{algorithm}

The binary tournament selection is performed using the crowding-comparison heuristic, while the replacement step is based on the non-dominated sorting heuristic (also the crowding-comparison in some cases). The non-dominated sorting results is partitioned in a set $F = \{F_1, \dots, F_K\}$ of $K$ non-dominated fronts of increasing rank $i$, where $i \in [1,K]$. Having $F_1$ the front that is not dominated by any other one, while the remaining fronts are dominated by all the ones that have a lower rank. On the basis of the crowding-comparison operator, the NSGA-II favours solutions of lower rank if the solutions being compared belong to different fronts. On the other hand, if the solutions come from the same front, it advantages the solution having a higher crowding distance. For more details about the non-dominated-sorting and crowding-comparison operators, one should refer to the NSGA-II original work \cite{DPA02}.

To solve the FL-COP, the NSGA-II is executed during a preliminary step in order to extract the optimal parameters of the \texttt{FedAvg} influencing the communication overhead. These parameters are: \texttt{(I)} the number $E$ of training steps performed before sending the local model to the server, \texttt{(II)} the number $m$ of clients participating in the training of the global model, \texttt{(III)} the number $b$ of bits used to encode the weights of each layer of the local model, and \texttt{(IV)} the percentage $\mu$ of weights that will not be transmitted. In the following, we provide more details about each of the NSGA-II steps when solving the FL-COP.

\vspace{-1em}

\subsubsection{Initialisation:}
\begin{figure}
    \centering
    \includegraphics[scale = 0.3]{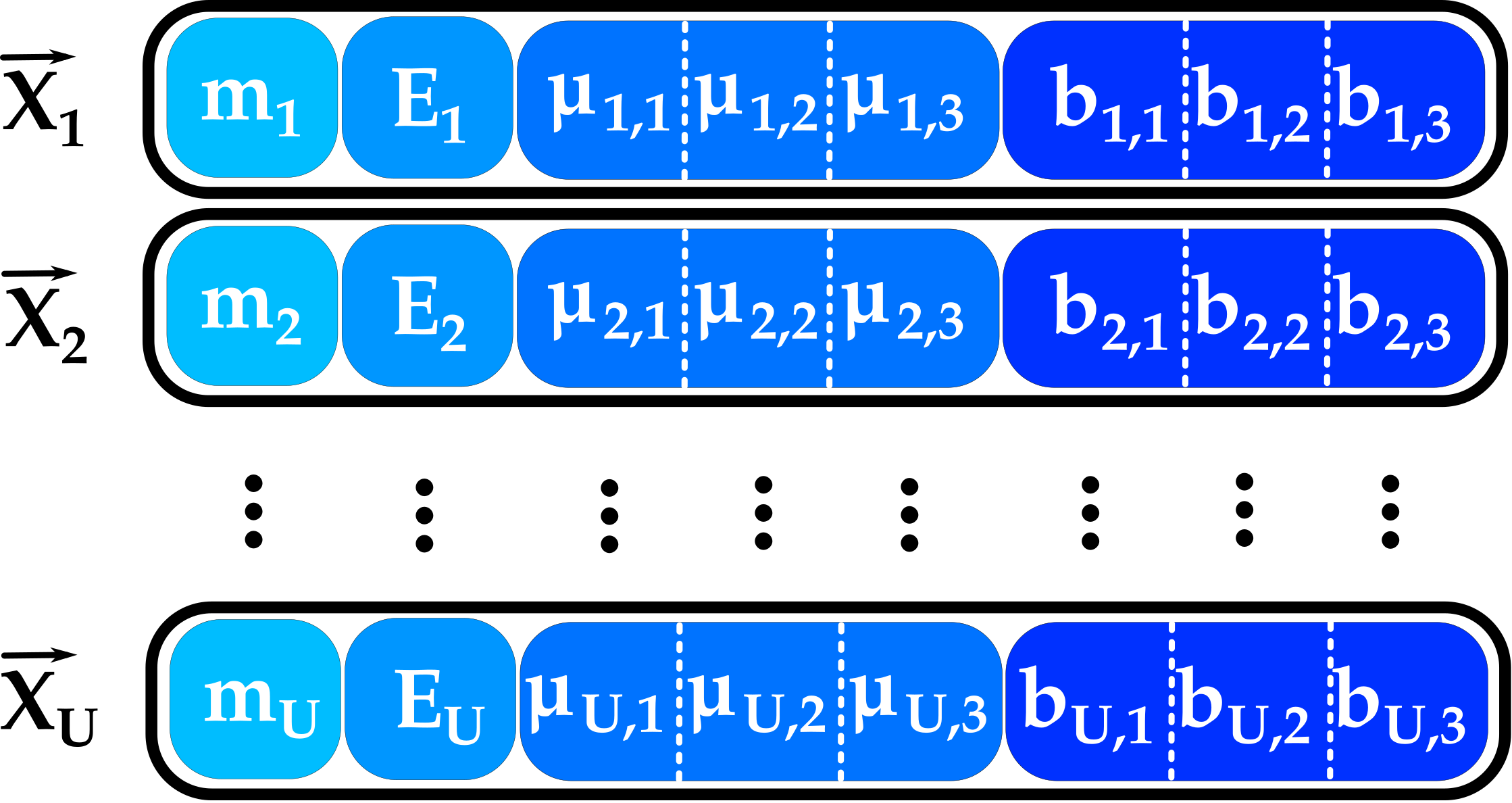}
    \caption{NSGA-II population for solving the FL-COP}
    \label{fig:nsgaflcop}
\end{figure}

The NSGA-II starts by initialising a population $P$ of $U$ solutions $\overrightarrow{X} = \{m,E,\mu_{1},\dots,\mu_{l},b_{1},\dots,b_{l}\}$ of size $d = (2 \cdot l + 2)$ knowing that $l$ is the number of layers in the trained model (see Fig. \ref{fig:nsgaflcop}). 

\vspace{-2.5em}

\subsubsection{Selection:}
As a second step, NSGA-II performs a binary tournament selection on the parent population $P$ to select the individuals that will undergo the breeding phases which will produce a population $Q$ of $U$ new offspring. The selection step creates a set $A$ of ($N/2$) pairs of parents, where the selection criterion is the crowding distance. 

\vspace{-1em}

\subsubsection{Crossover:}

\begin{figure}
    \centering
    \includegraphics[scale = 0.25]{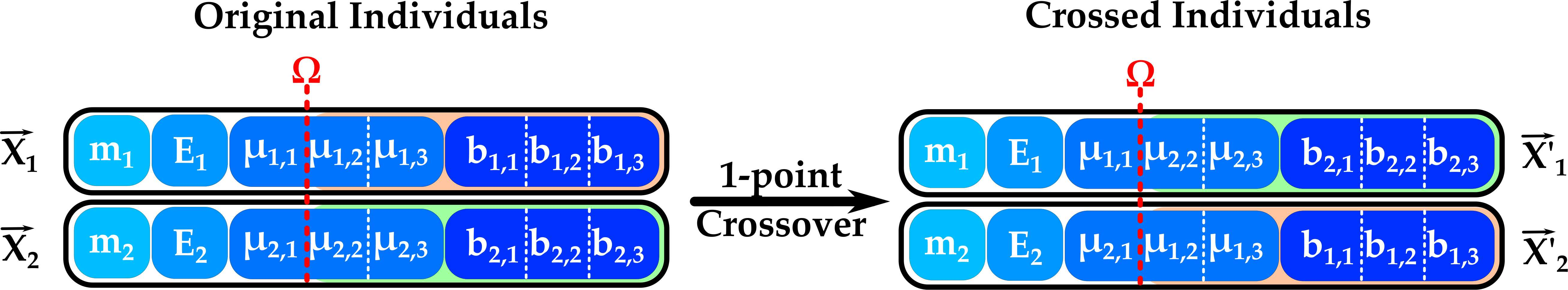}
    \caption{Single-point crossover applied on FL-COP solutions }
    \label{fig:spx}
\end{figure}

Afterwards, according to a probability $p_{c}$, each pair of parents from the set $A$ of selected ones will (or not) undergo the single-point crossover. Our aim is to prove that even using relatively-simple operators, the NSGA-II can still solve the FL-COP adequately. The crossover step will result in a new set $B$ of $U$ crossed offspring. It randomly chooses a switching point $\Omega$ from the interval [1,$d$] and exchanges the solutions' substrings delimited by the variables at the position $\Omega$ and $d$. Fig. \ref{fig:spx} illustrates a single-point crossover applied on two individuals $\overrightarrow{X_{1}}$ and $\overrightarrow{X_{2}}$, representing a solution for FL-COP with a 3-layers model. Once applied, the crossover results in two new offspring $\overrightarrow{X'_{1}}$ and $\overrightarrow{X'_{2}}$.

\vspace{-1em}

\vspace{-2.5em}

\subsubsection{Mutation:}

During this step, each offspring in the set $B$ obtained during the crossover phase will undergo (or not) a uniform mutation that is ruled by a probability $p_{m}$. The mutation phase results in a new population $Q$ of $U$ offspring. Like the crossover, our goal is to prove that even using operators of low complexity, the NSGA-II can still provide a meaningful efficiency. \\

\begin{figure}
    \centering
    \includegraphics[scale = 0.25]{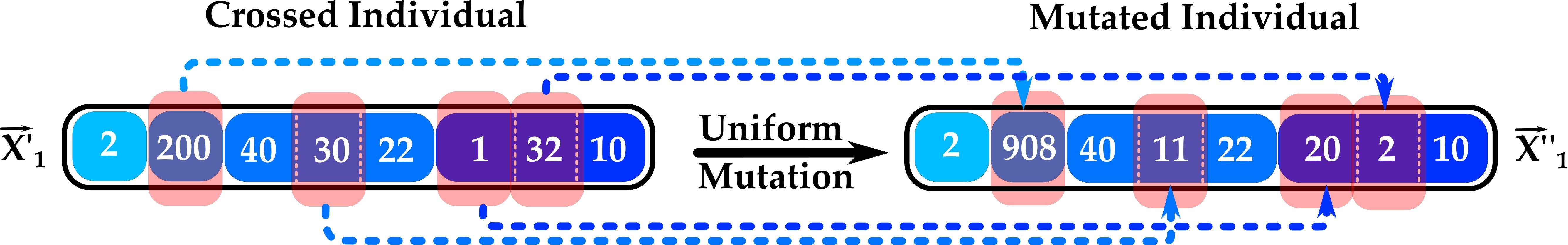}
    \caption{Uniform mutation applied on FL-COP solutions}
    \label{fig:unfimut}
\end{figure}

The uniform mutation generates, for each variable of the solution being mutated, a random number from the interval $[\tau, \varrho]$, where $\tau$ and $\varrho$ are the upper and lower bounds in which the variable being mutated can take valid values:  $E \in [1,100]$, $m \in [1,4]$, $\mu \in [0,50]$ and $b \in [1,32]$. Fig. \ref{fig:unfimut} illustrates an example of the uniform mutation applied to an individual $\overrightarrow{X'_{1}}$ resulting from the crossover and producing a mutated individual $\overrightarrow{X''_{1}}$. The FL-COP in this case concerns a 3-layers mode, where the original individual illustrates a 200-iterations training, transfers 70$\%$ of the 2$^{nd}$ layer's weights and encodes the weights of 1$^{st}$ and the 2${^{nd}}$ layers using 1 and 32 bits. Once mutated, the individual represents a 908-iterations training, will send 89$\%$ of the 2$^{nd}$ layer's weights and finally the weights of 1$^{st}$ and the 2${^{nd}}$ layers will be encoded using 20 and 2 bits, respectively. 

\vspace{-.5em}

\vspace{-2.5em}

\subsubsection{Replacement:}

Having the original population $P$ as well as $Q$ obtained after applying selection, crossover and mutation, a replacement step is applied in order to decide the composition of the population $P'$ during the next iteration. Having $F = \{F_{1},\ldots,  F_{K}\}$ the set of non-dominated fronts obtained after applying the non-dominated sorting heuristic, $P'$ will be filled by including the non-dominated fronts in an increasing rank until the $|P'| = N$. Let us admit that at some moment, one wants to include the $i^{th}$ front, but the union of both $P'$ and $F_{i}$ is greater than $N$. In this case, the solutions of the $i^{th}$ front are sorted in a descending order based on the crowding-comparison operator. Then, this $P'$ will be filled by including the missing solutions from the best ones obtained after ranking the $i^{th}$ front.

\vspace{-1em}
\section{Experimental Study and Analysis \label{sec:esa}}
In this section, we provide details of the experiments conducted to assess our proposal, as well as the obtained results and their discussion.

\vspace{-1em}

\subsection{Problem Benchmarks and Experimental Settings}
\vspace{-0.5em}

The implementation\footnote{\url{https://github.com/NEO-Research-Group/flcop}} has been made in \texttt{Python} version 3.8.8, while the execution has been done in the Picasso supercomputing center at the University of Malaga. In particular, we used two types of hardware from a computation cluster: \texttt{(I)} a 24 $\times$ Bull R282-Z90 nodes: 128 cores (AMD EPYC 7H12 @ 2.6GHz), 2 TB of RAM, Infini-Band HDR200 network, 3.5 TB of local-scratch disks. \texttt{(II)} a 4 $\times$ DGX-A100 nodes: 8 GPUs (A100 Tensor Core), 1 TB of RAM, Infini-Band FDR40 network and a 14 TB of local-scratch. We use a process with 128 cores and 400 GB of RAM to evaluate the solutions in parallel.

Our experiments have been thought of to assess our proposal's solving efficiency and scalability when dealing with different sizes of FL-COP benchmarks and its adaptability when dealing with different types of neural network models. Thus, we consider both convolutional and fully-connected Neural Network (NN) topologies with 12 and 3 layers and 887,530 and 33,400  weights, respectively. Our experiments have been done using the well-established \texttt{MNIST} dataset containing 70,000 images of handwritten digits. At the beginning of each execution, the initial weights of the models are drawn using the same seed. NSGA-II has been run using a population of 100 individuals, $p_{c}$ = 0.9 and $p_{m}$ = (1/$d$). The size $d$ of the solutions in the case of the convolutional model is 26, while in the fully-connected, it is 8. NSGA-II has been executed 30 times for each NN type, where it is executed during 300 iterations for the fully-connected model and 120 on the convolutional one. 

The experimentation has been done by randomly distributing 60,000 images of the \texttt{MNIST} training set among the $m$ clients. For simplicity, all clients have the same number of data. Each partition of the data remains private in each client throughout the learning process. We evaluate the final models obtained with the 10,000 images of the \texttt{MNIST} dataset. We conduct two different types of experiments. First, we apply NSGA-II for solving the FL-COP that uses a fully-connected neural network with 33,400 trainable parameters and 3 layers: one input (784), one intermediate (42), and one output layer (10). The middle layer and the output layer have a bias. We consider each bias of each layer as an independent array to optimise. Therefore, we have an array of weights of length 4 (i.e. [32928, 42, 420, 10]). In our experiments, we fix $l = 4$ where the two additional layers are of the bias. In the second experiment, we do the same on a convolutional neural network with 887,530 trainable parameters. In this case, we have a multidimensional array of length 12 (i.e. [800, 32, 25600, 32, 18432, 64, 36864, 64, 802816, 256, 2560, 10]). The termination criterion in both experiments is achieving one epoch (i.e. all clients trained with all their local data one time). In this experiment, we fix $l = 12$. In both cases, we simulate a server/client architecture of 4 clients (i.e. $N$ = 4). All our results have been confirmed using a Wilcoxon test with Bonferroni correction and a significance level of 0.025.

\vspace{-1.1em}
\subsection{Experimental Results and Discussion }
\begin{figure}[!htb]
   \begin{minipage}{0.48\textwidth}
     \centering
     \includegraphics[width=.95\linewidth]{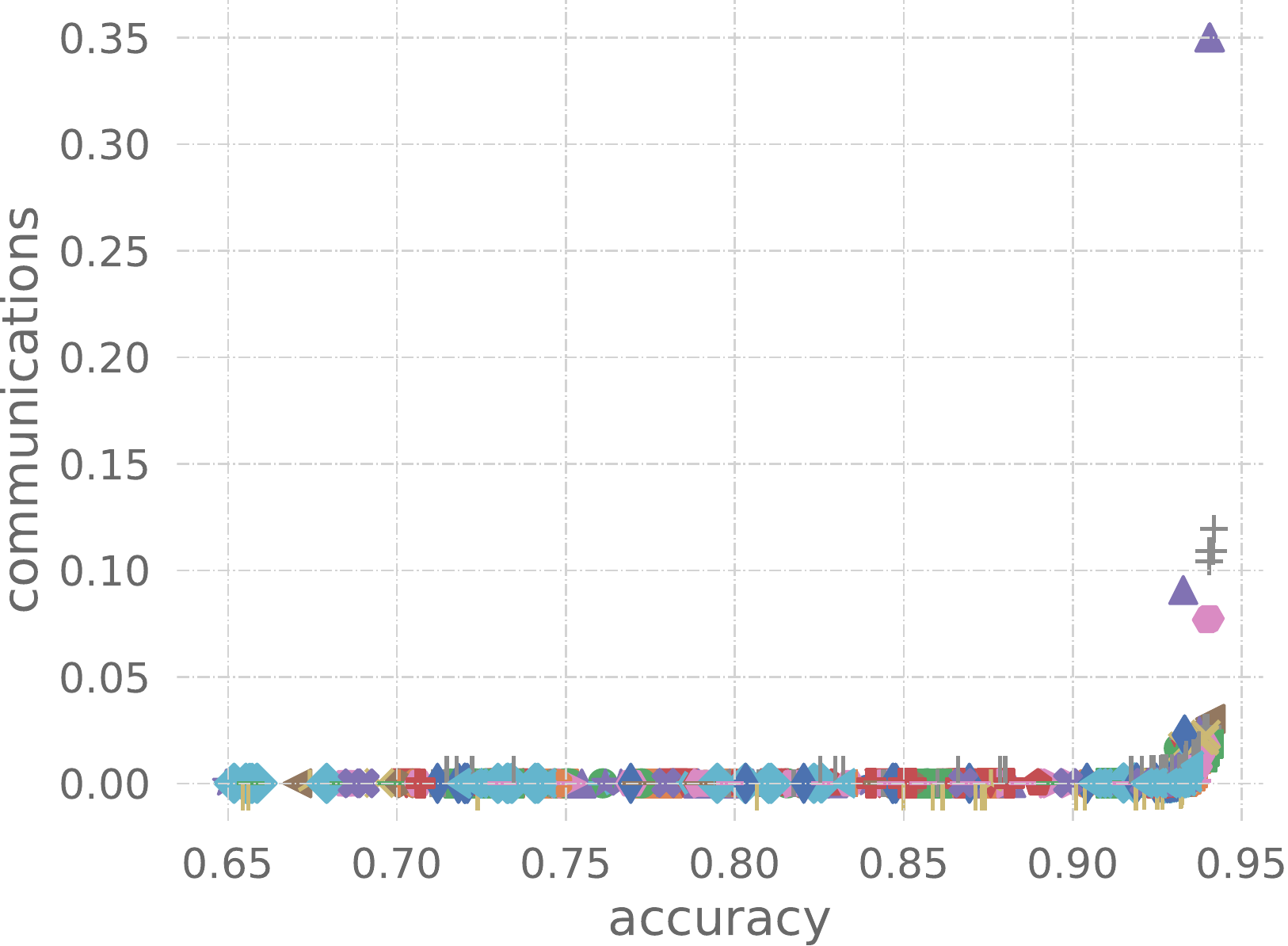}
     \caption{Fully-connected NN: 30 executions' Pareto fronts (1 color/Pareto).}\label{Fig:dense1}
   \end{minipage}\hfill
   \begin{minipage}{0.48\textwidth}
     \centering
     \includegraphics[width=.95\linewidth]{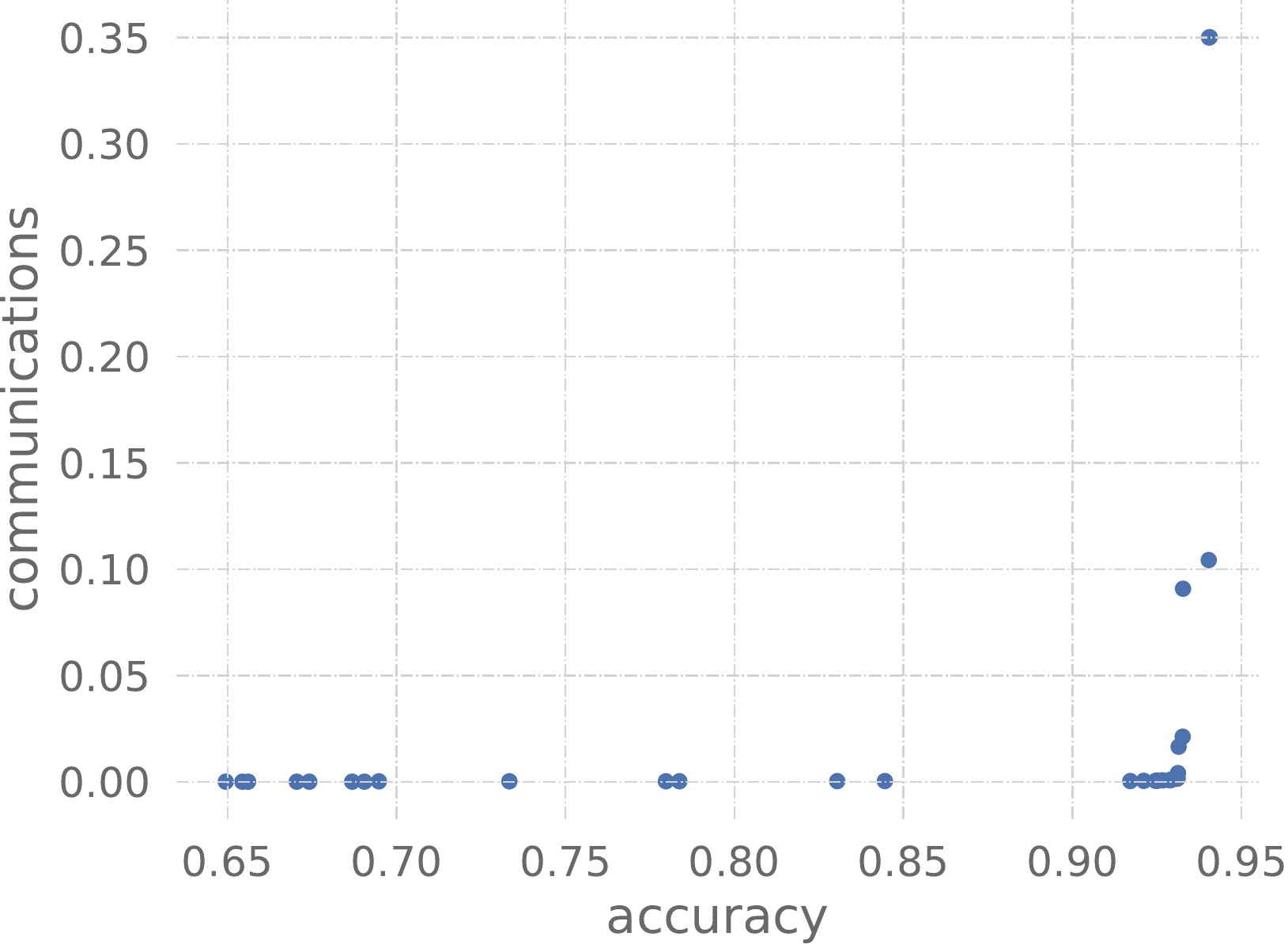}
     \caption{Fully-connected NN: pseudo-optimal Pareto front.}\label{Fig:dense2}
   \end{minipage}
\end{figure}

\begin{figure}[!htb]
   \begin{minipage}{0.48\textwidth}
     \centering
     \includegraphics[width=0.95\linewidth]{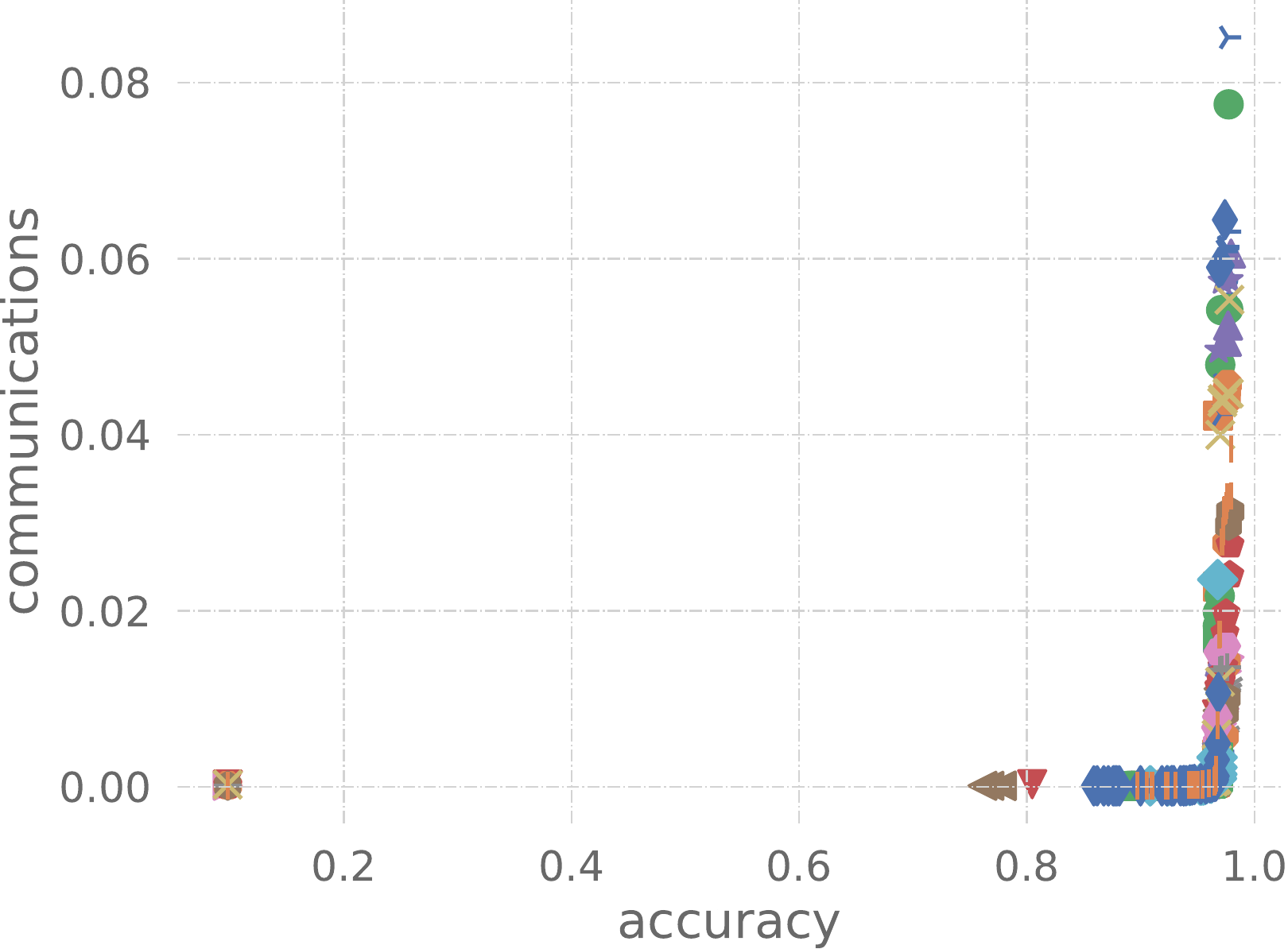}
     \caption{Convolutional NN: 30 executions' Pareto fronts (1 color/Pareto).}\label{Fig:conv1}
   \end{minipage}\hfill
   \begin{minipage}{0.48\textwidth}
     \centering
     \includegraphics[width=0.95\linewidth]{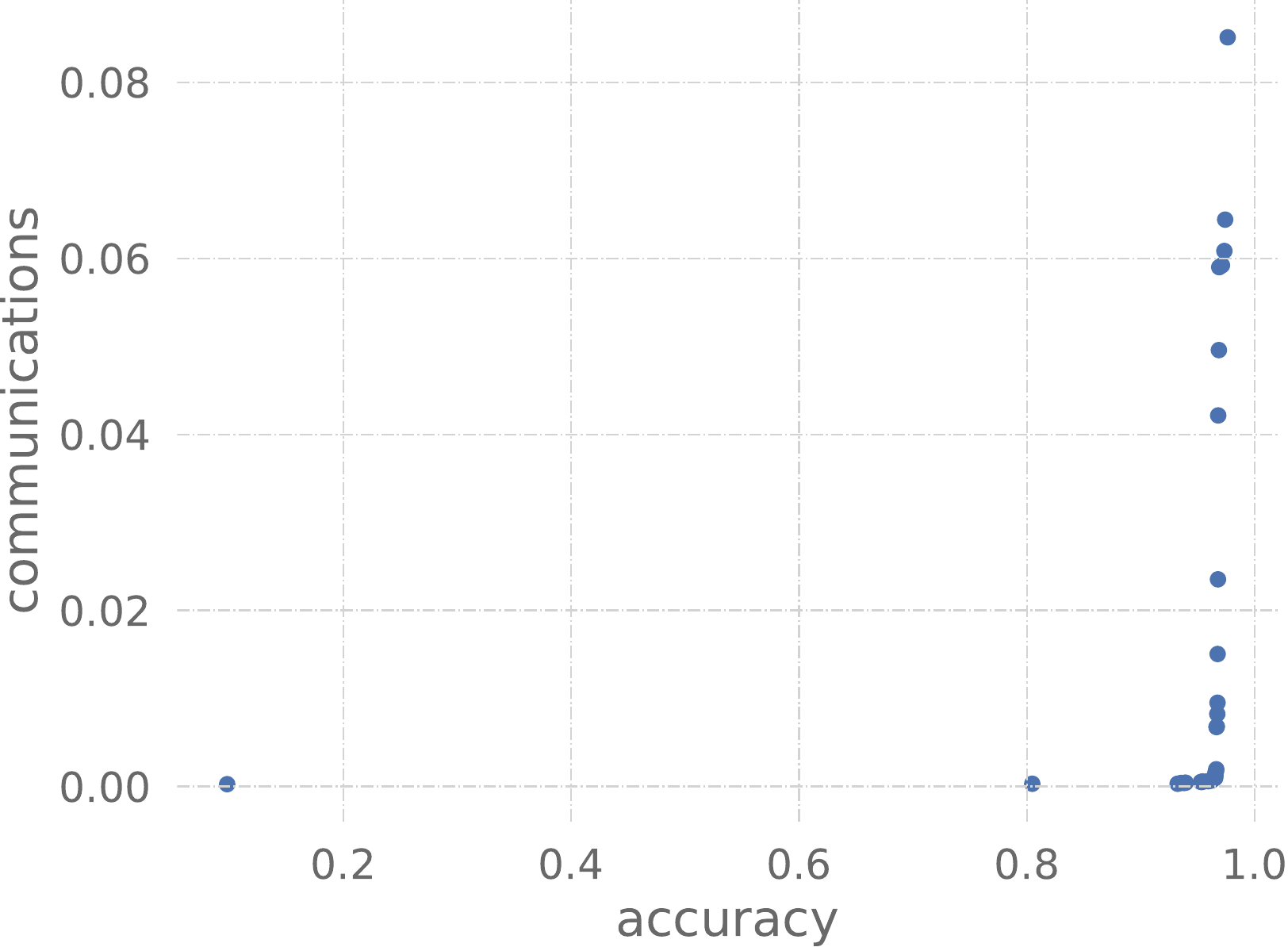}
     \caption{Convolutional NN: pseudo-optimal Pareto front.}\label{Fig:conv2}
   \end{minipage}
\end{figure}

\vspace{-.4em}
Considering the fully-connected NN topology, Fig. \ref{Fig:dense1} illustrates the Pareto fronts obtained by NSGA-II in 30 executions, while Fig. \ref{Fig:dense2} represents the pseudo-optimal Pareto front that dominates all those obtained in 30 runs. Saying that for the convolutional we executed 25 executions. Similarly, Figs.~\ref{Fig:conv1} and~\ref{Fig:conv2} present the same information for the convolutional model. It can be seen in Fig.~\ref{Fig:dense1} and~\ref{Fig:dense2} that our proposal could reduce the communication to 35\% in the worst solution, and to nearly 0\% of communication, while maintaining accuracy above 0.94. Considering, Figs.~\ref{Fig:conv1} and~\ref{Fig:conv2}, one can note that our approach could reduce the communication to 6\% in the worst case, while it could achieve nearly 0\% communication with an accuracy above 0.95.

\vspace{-1.25em}

\vspace{-2.em}

%\vspace{-2em}

\begin{figure}[!htb]
   \begin{minipage}{0.48\textwidth}
     \centering
     \includegraphics[width=1.\linewidth]{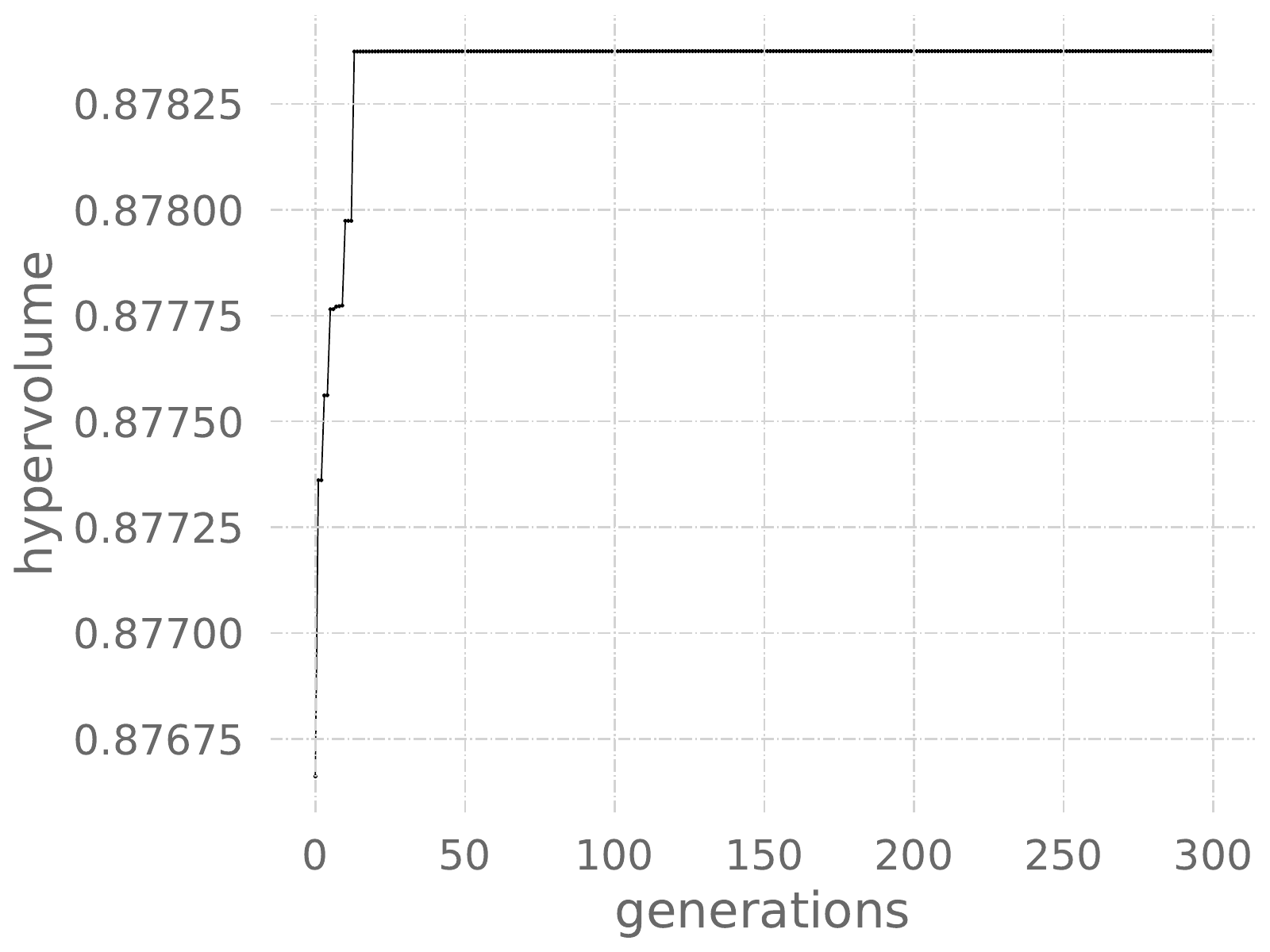}
     \caption{Fully-connected NN: Hypervolume evolution through iterations.}\label{Fig:hypv1}
   \end{minipage}\hfill
   \begin{minipage}{0.48\textwidth}
     \centering
     \includegraphics[width=1.\linewidth]{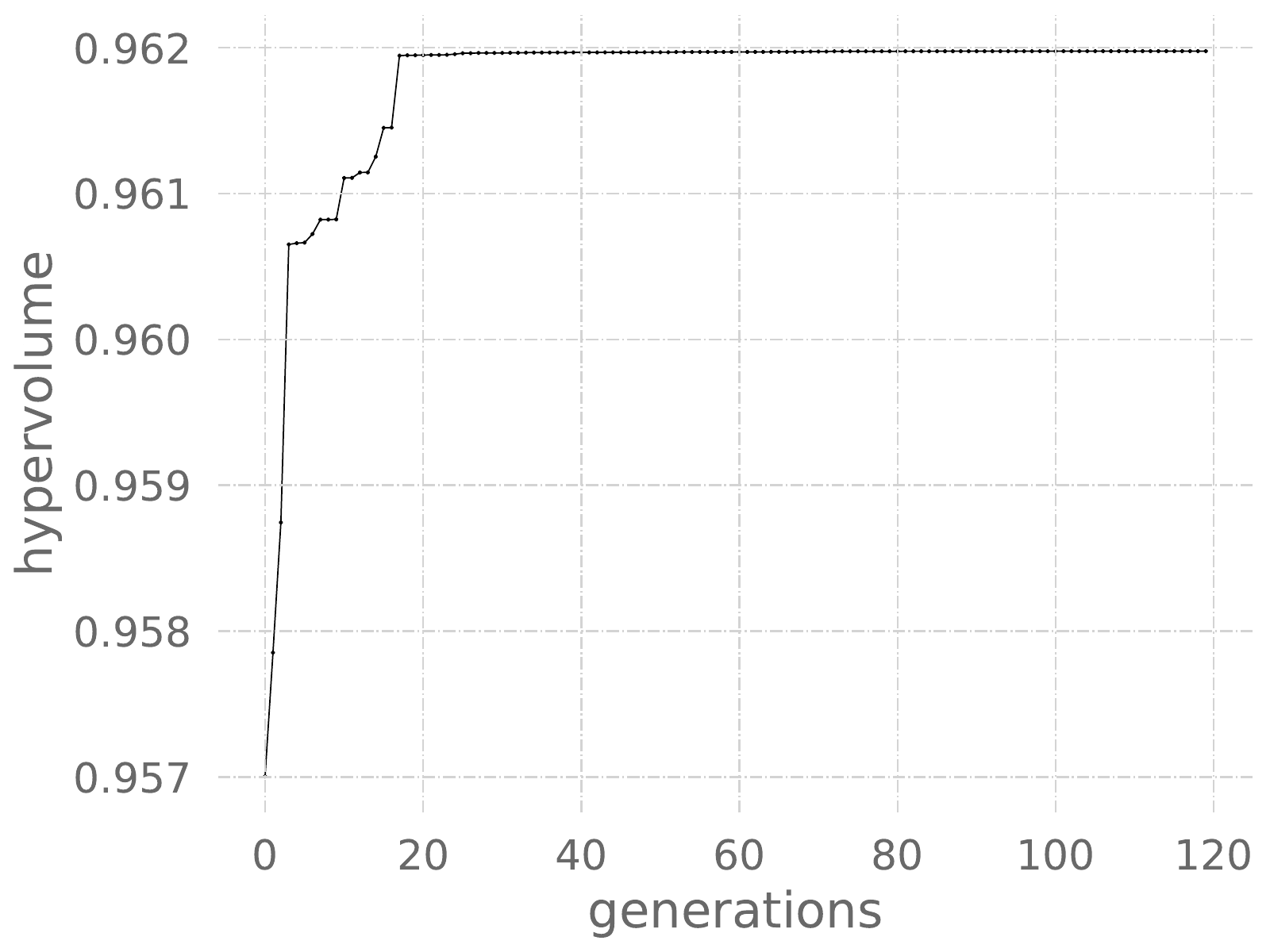}
     \caption{Convolutional NN: Hypervolume evolution through iterations.}\label{Fig:hypv2}
   \end{minipage}
\end{figure}

Figs.~\ref{Fig:hypv1} and~\ref{Fig:hypv2} illustrate the evolution of the average hypervolume of the Pareto fronts obtained by NSGA-II throughout one randomly selected, but yet representative execution when tackling the FL-COP using a fully-connected and convolutional neural network, respectively. The smooth evolution of hypervolume in the first iterations can be explained by the fact that NSGA-II starts with random low-quality individuals that can be quickly enhanced. Nonetheless, as the iterations go, the attained Paretos are of higher quality and difficult to enhance beyond iteration 40. Of course, more advanced hypotheses could be made to explain such behaviour, nonetheless it will be hard to confirm them without further in-depth analysis.

%\vspace{-1.5em}

\iffalse
%Higher ACC
%Mean, max,min, std ACC
0.9351899981498718
0.9416999816894531
0.9314000010490417
0.0032892092778994114
%Mean, max,min, std COMM
0.028198848764057892
0.35011385826120955
0.001978788982680557
0.06426531563959363

%Pendiente
%Mean, max,min, std ACC
0.9348399976889292
0.9408000111579895
0.9314000010490417
0.002998733798142603
%Mean, max,min, std COMM
0.016434431594654186
0.0908464960450357
0.001978788982680557
0.019718137459287466

%Lower Comm
%Mean, max,min, std ACC
0.700623337427775
0.7251999974250793
0.6495000123977661
0.02312765972478509
0.00013709298084825609
%Mean, max,min, std COMM
0.00029205667619824904
0.00012695312499999556
2.9617630999786965e-05
\fi

\begin{sidewaysfigure}
     \centering
     \begin{subfigure}[b]{0.3\textwidth}
         \centering
         \includegraphics[width=0.9\textwidth]{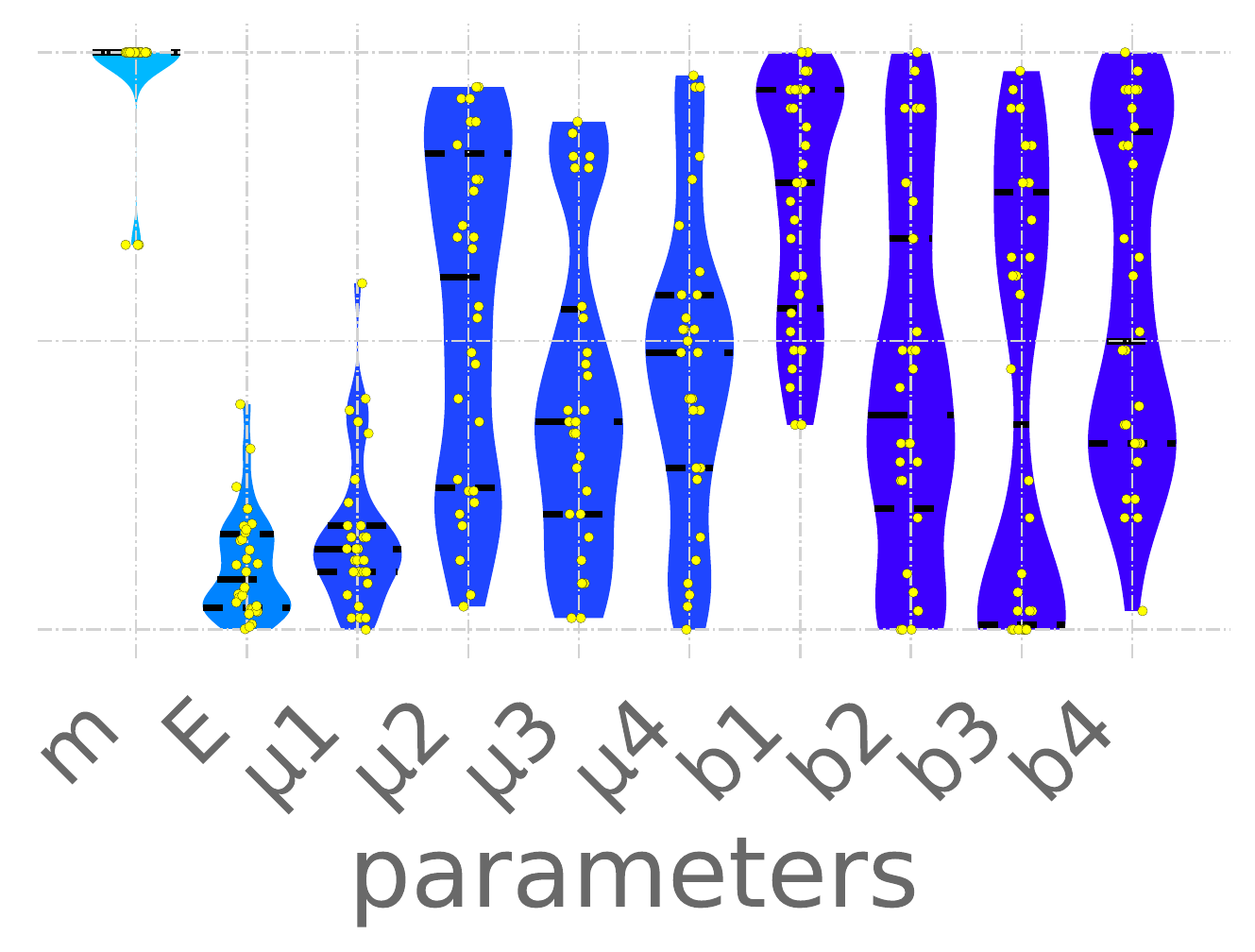}
         \caption{}
         \label{fig:params_1_dense}
     \end{subfigure}
     \hfill
     \begin{subfigure}[b]{0.3\textwidth}
         \centering
         \includegraphics[width=0.9\textwidth]{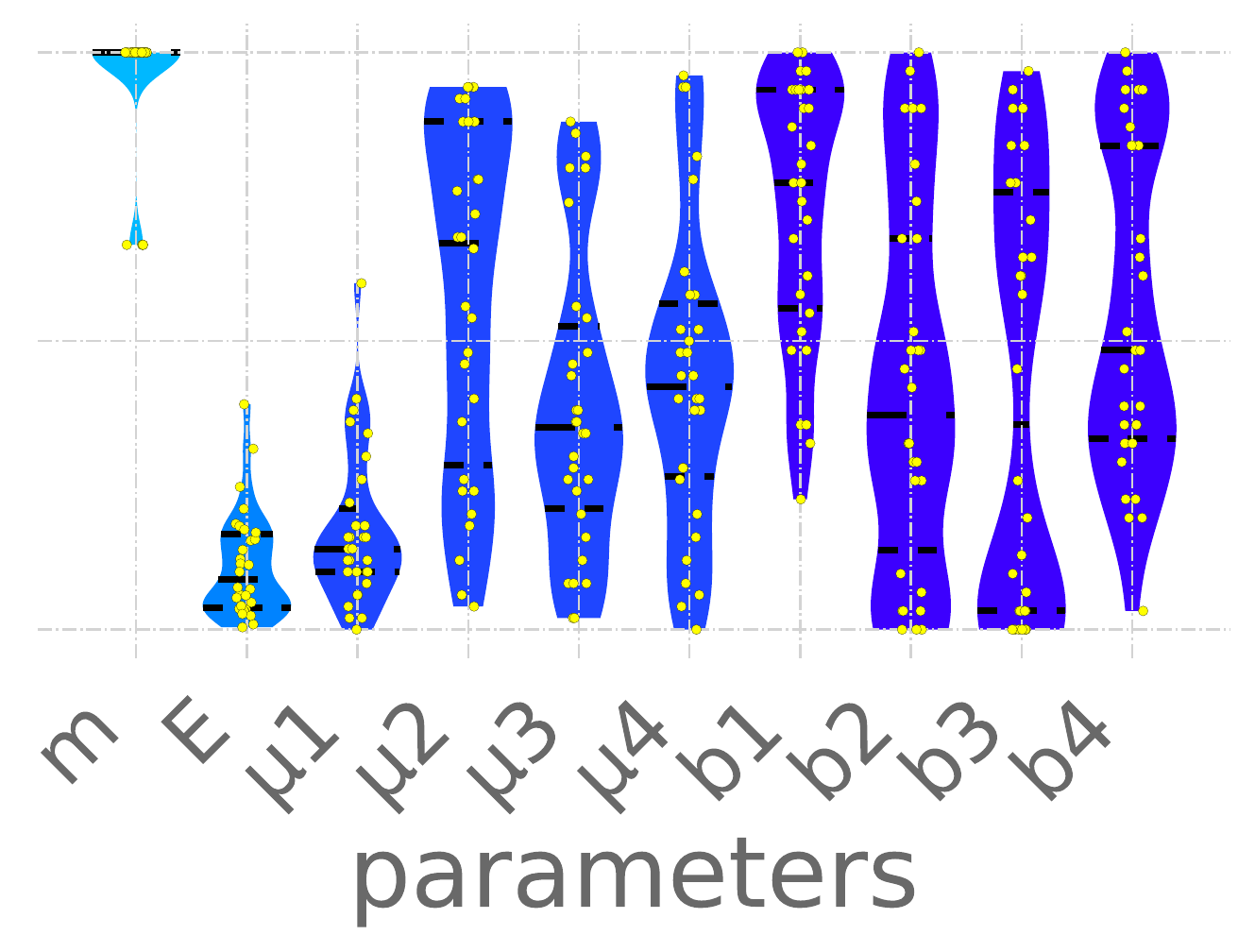}
         \caption{}
         \label{fig:params_2_dense}
     \end{subfigure}
     \hfill
     \begin{subfigure}[b]{0.3\textwidth}
         \centering
         \includegraphics[width=0.9\textwidth]{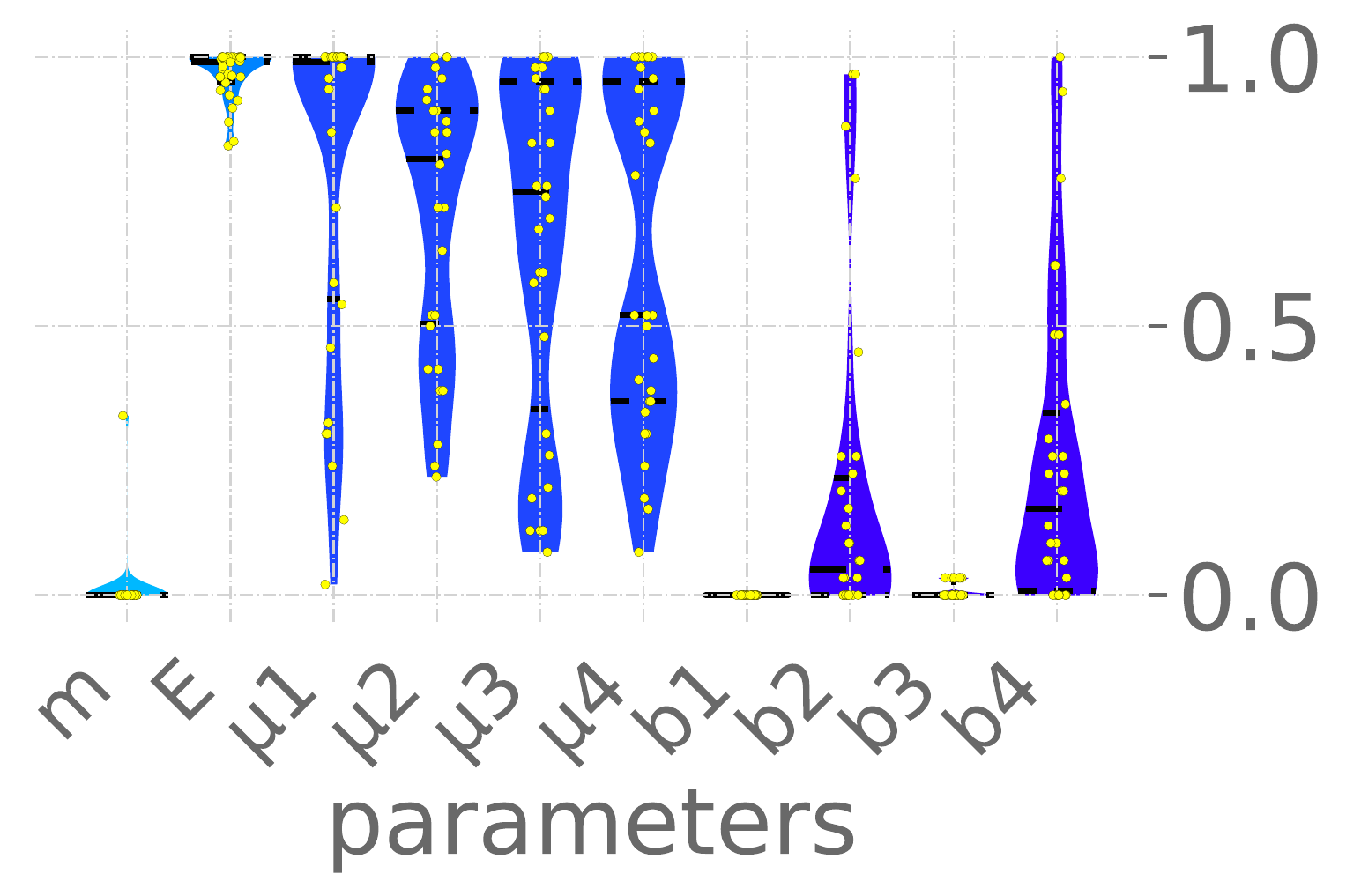}
         \caption{}
         \label{fig:params_3_dense}
     \end{subfigure}
     \hfill
     \vspace{-0.5em}
     \begin{subfigure}[b]{0.32\textwidth}
         \centering
         \includegraphics[width=\textwidth]{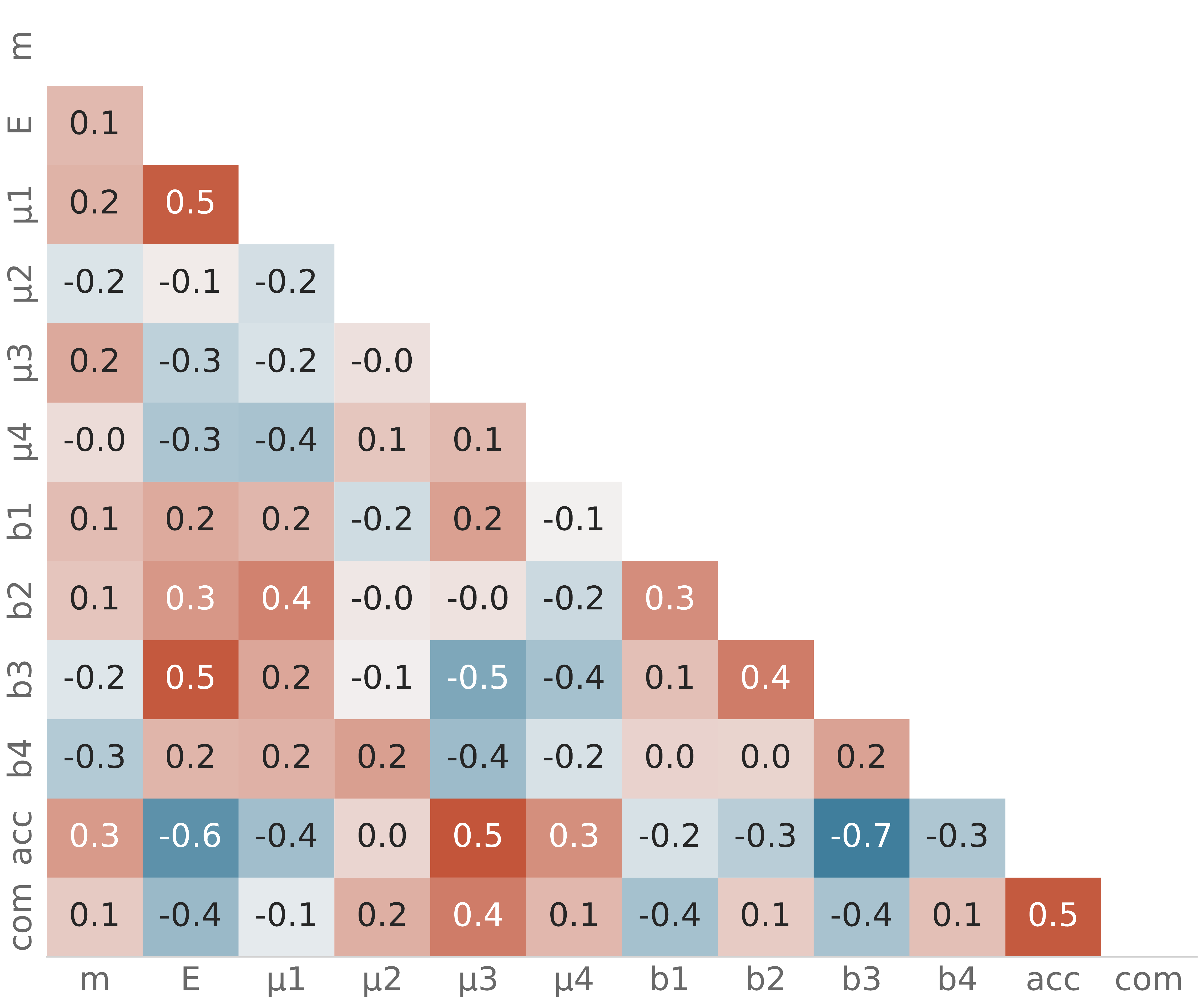}
         \caption{}
         \label{fig:heat_1_dense}
     \end{subfigure}
     \hfill
     \begin{subfigure}[b]{0.32\textwidth}
         \centering
         \includegraphics[width=\textwidth]{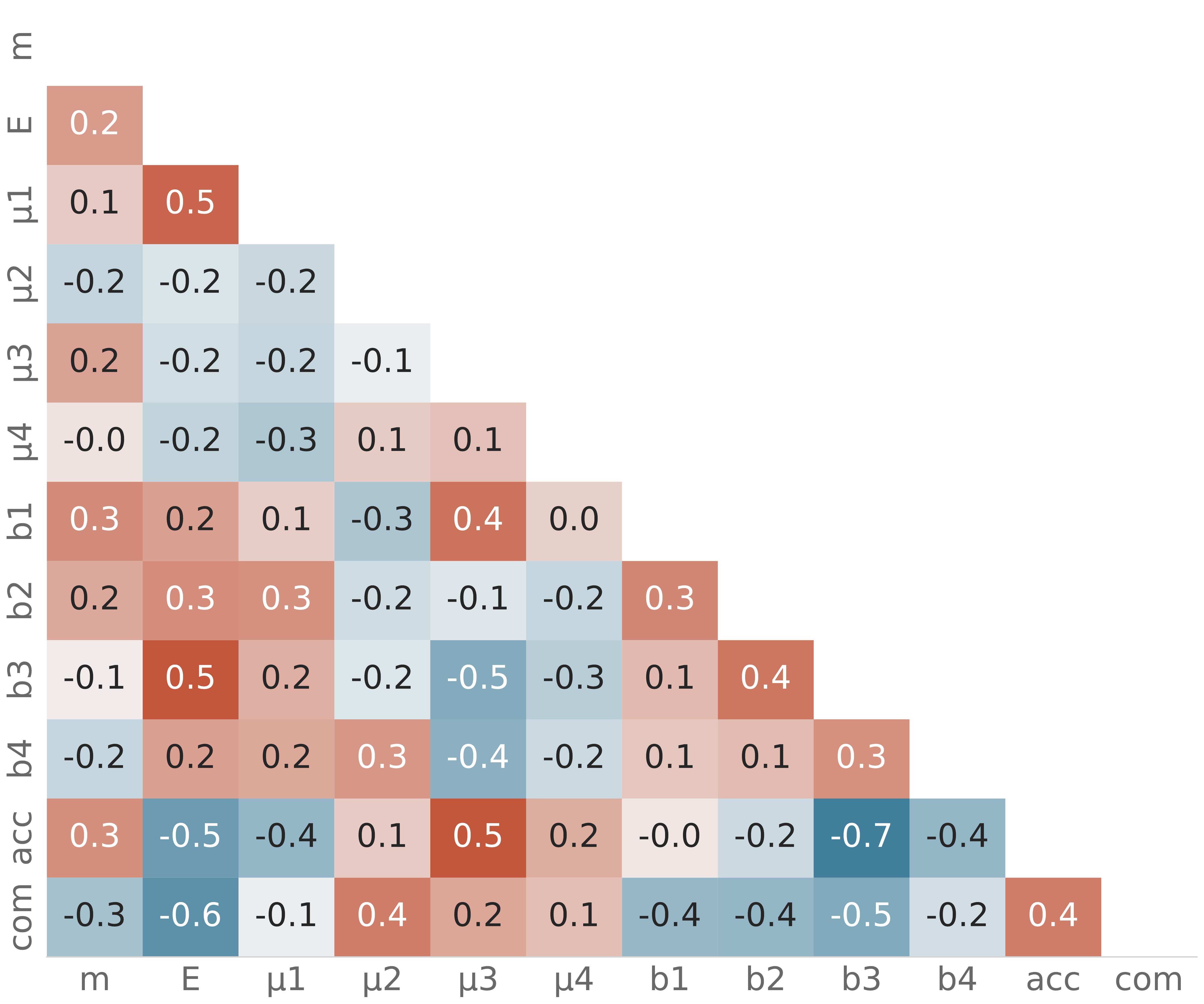}
         \caption{}
         \label{fig:heat_2_dense}
     \end{subfigure}
     \hfill
     \begin{subfigure}[b]{0.32\textwidth}
         \centering
         \includegraphics[width=\textwidth]{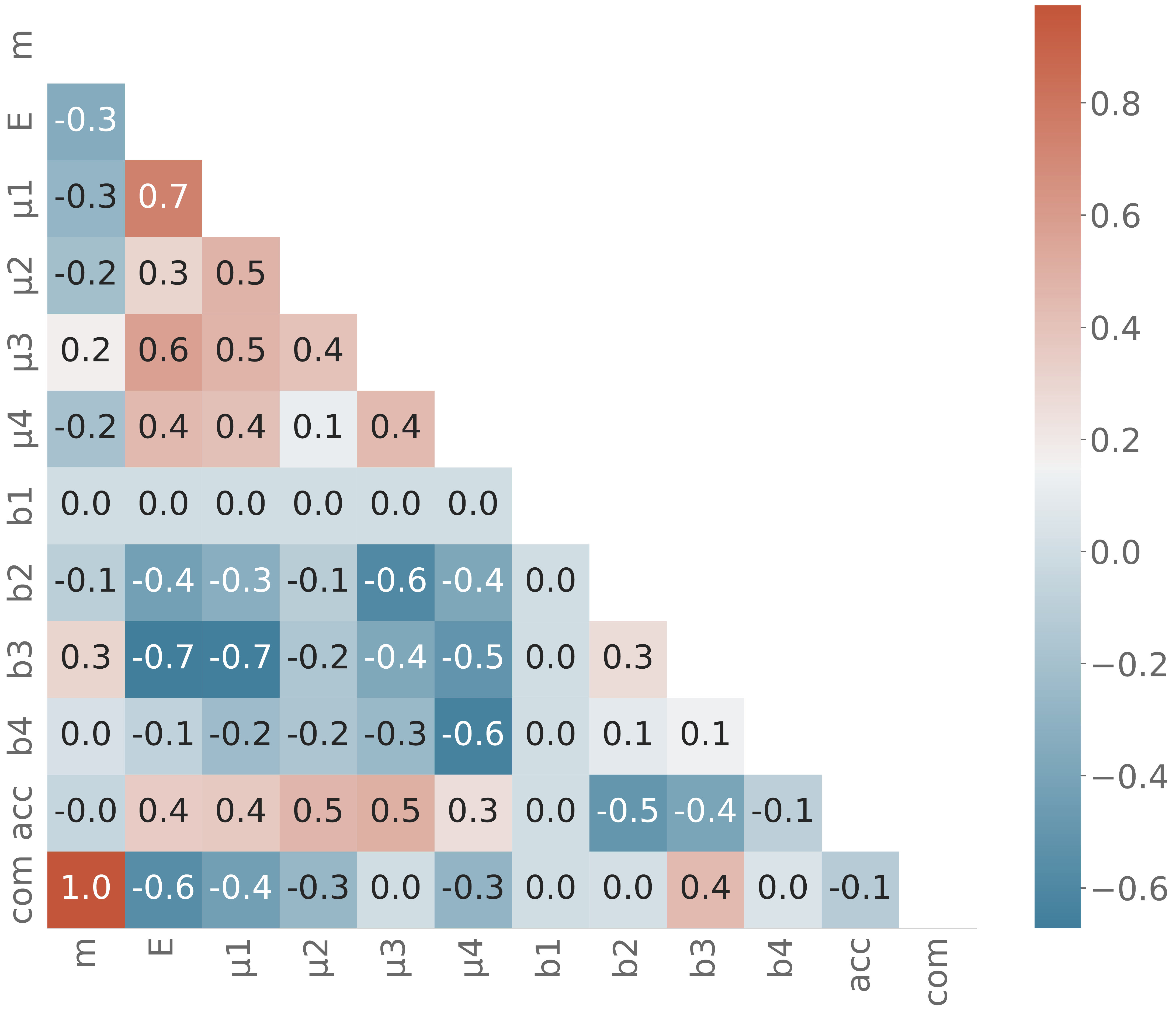}
         \caption{}
         \label{fig:heat_dense}
     \end{subfigure}  
     \hfill

     \begin{subfigure}[b]{0.3\textwidth}
         \centering
         \includegraphics[width=0.8\textwidth]{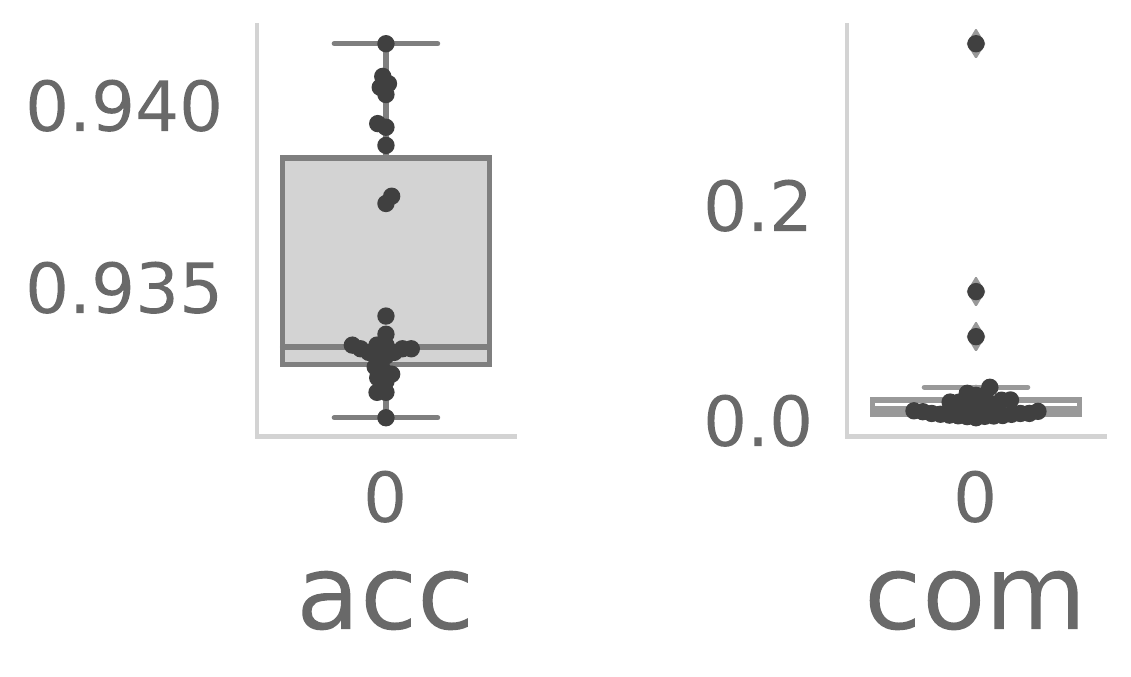}
         \caption{}
         \label{fig:heat_1_dense}
     \end{subfigure}
     \hfill
     \begin{subfigure}[b]{0.3\textwidth}
         \centering
         \includegraphics[width=0.8\textwidth]{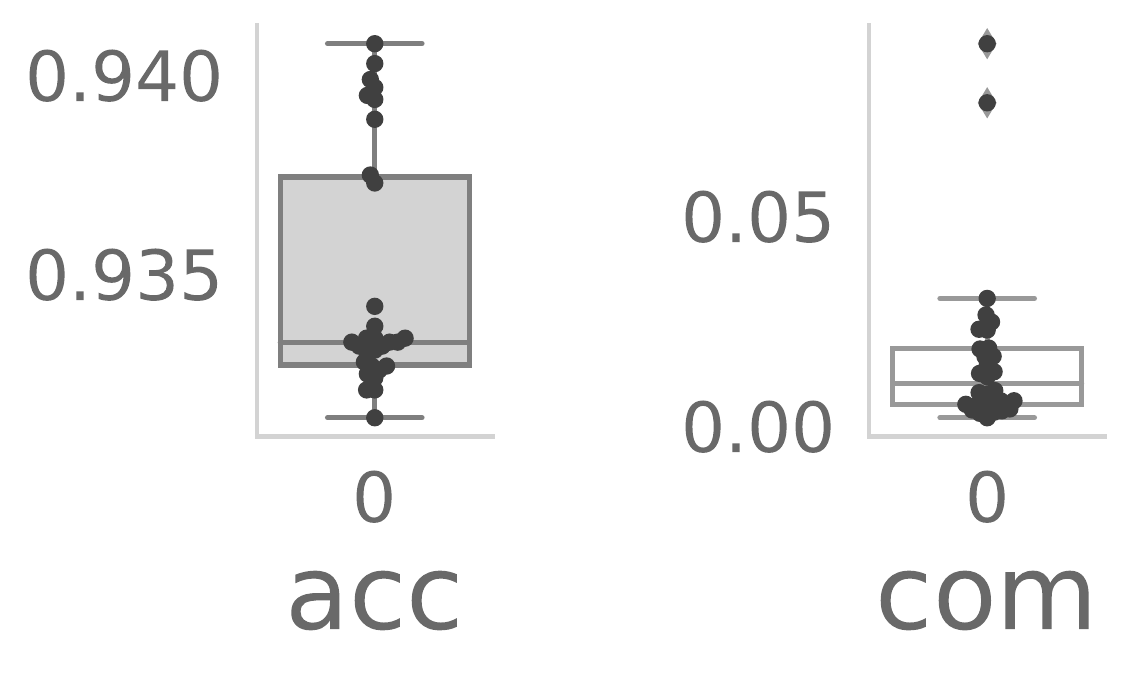}
         \caption{}
         \label{fig:heat_2_dense}
     \end{subfigure}
     \hfill
     \begin{subfigure}[b]{0.3\textwidth}
         \centering
         \includegraphics[width=0.8\textwidth]{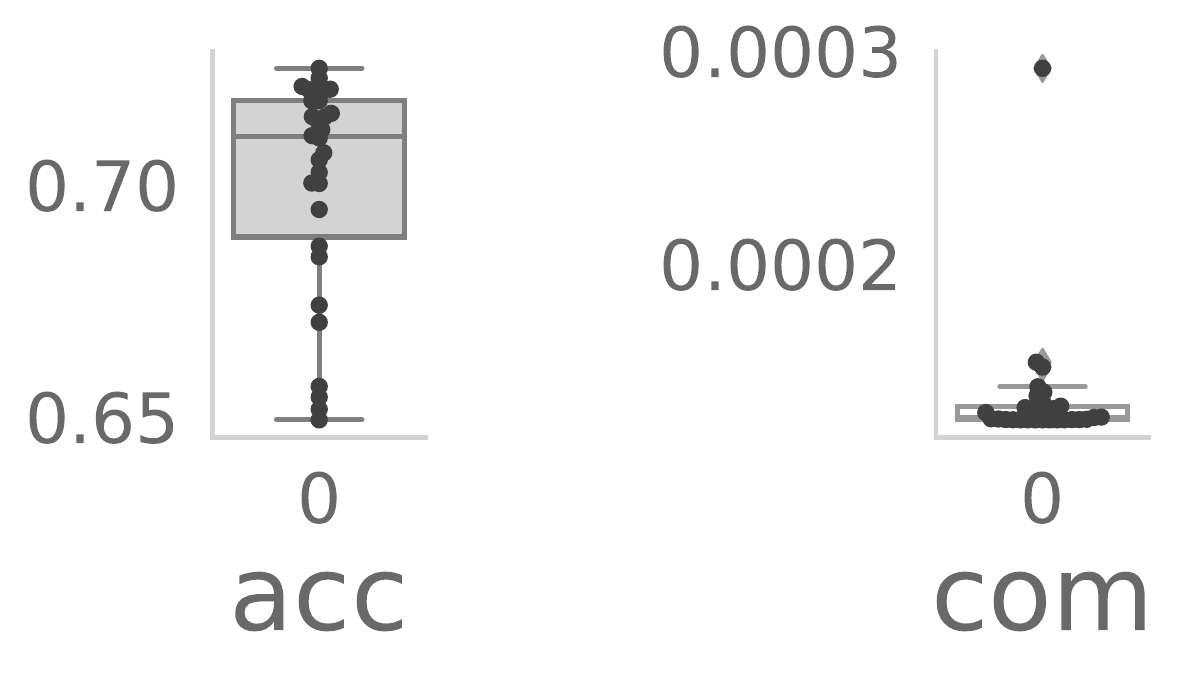}
         \caption{}
         \label{fig:heat_dense}
     \end{subfigure}  
     \vspace{-1em}
        \caption{Fully-connect NN: (a)-(c) Normalised parameters of the individuals with higher accuracy in final Pareto fronts. (d)-(f) Normalised parameters of the individuals with best correlation accuracy vs. communications in final Pareto fronts. (g)-(i) Accuracy and communications mean, std, max, min and quartiles.}
         \label{fig:dense_stats}
\end{sidewaysfigure}

\begin{sidewaysfigure}
     \centering
     \begin{subfigure}[b]{0.32\textwidth}
         \centering
         \includegraphics[width=\textwidth]{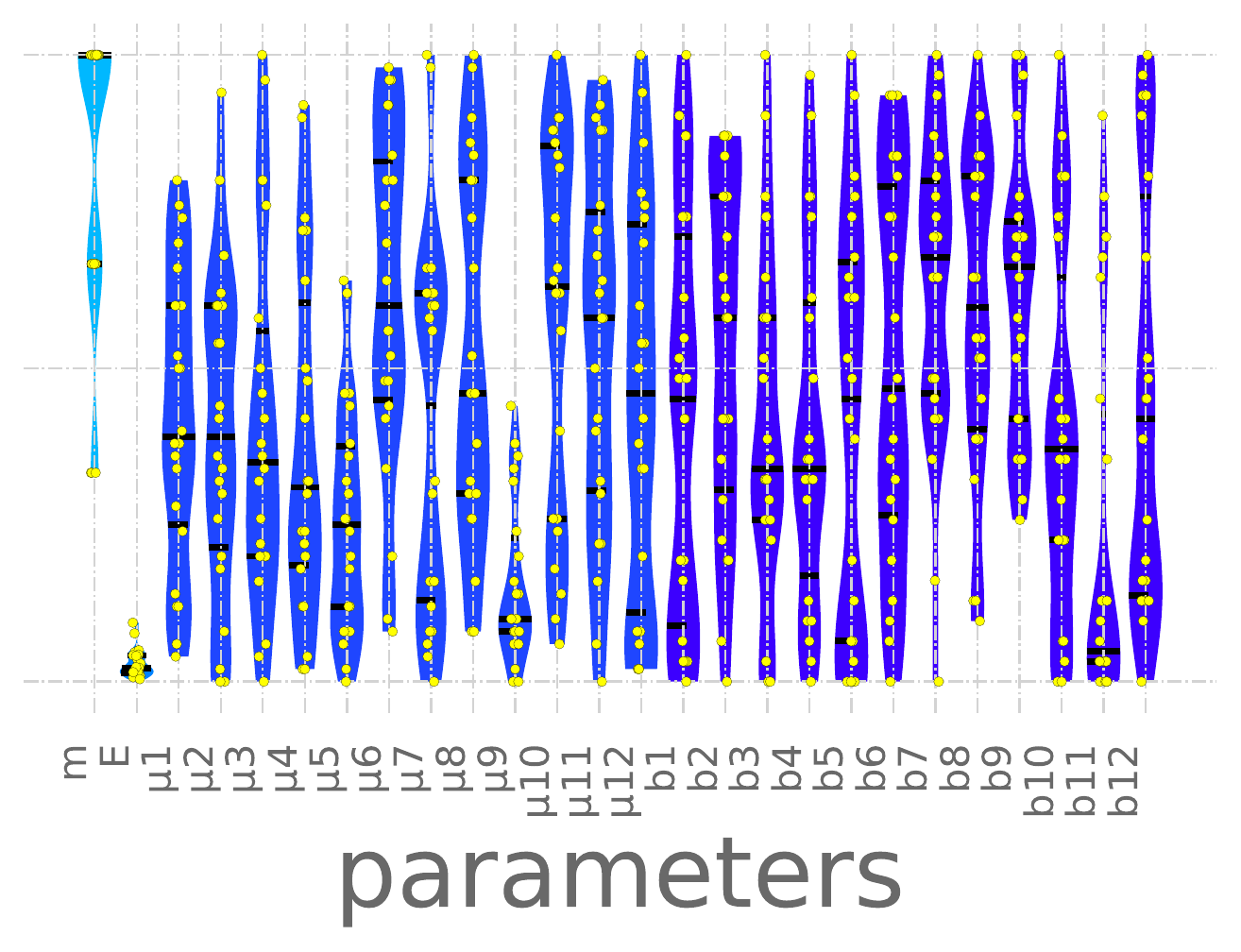}
         \caption{}
         \label{fig:params_1_dense}
     \end{subfigure}
     \hfill
     \begin{subfigure}[b]{0.32\textwidth}
         \centering
         \includegraphics[width=\textwidth]{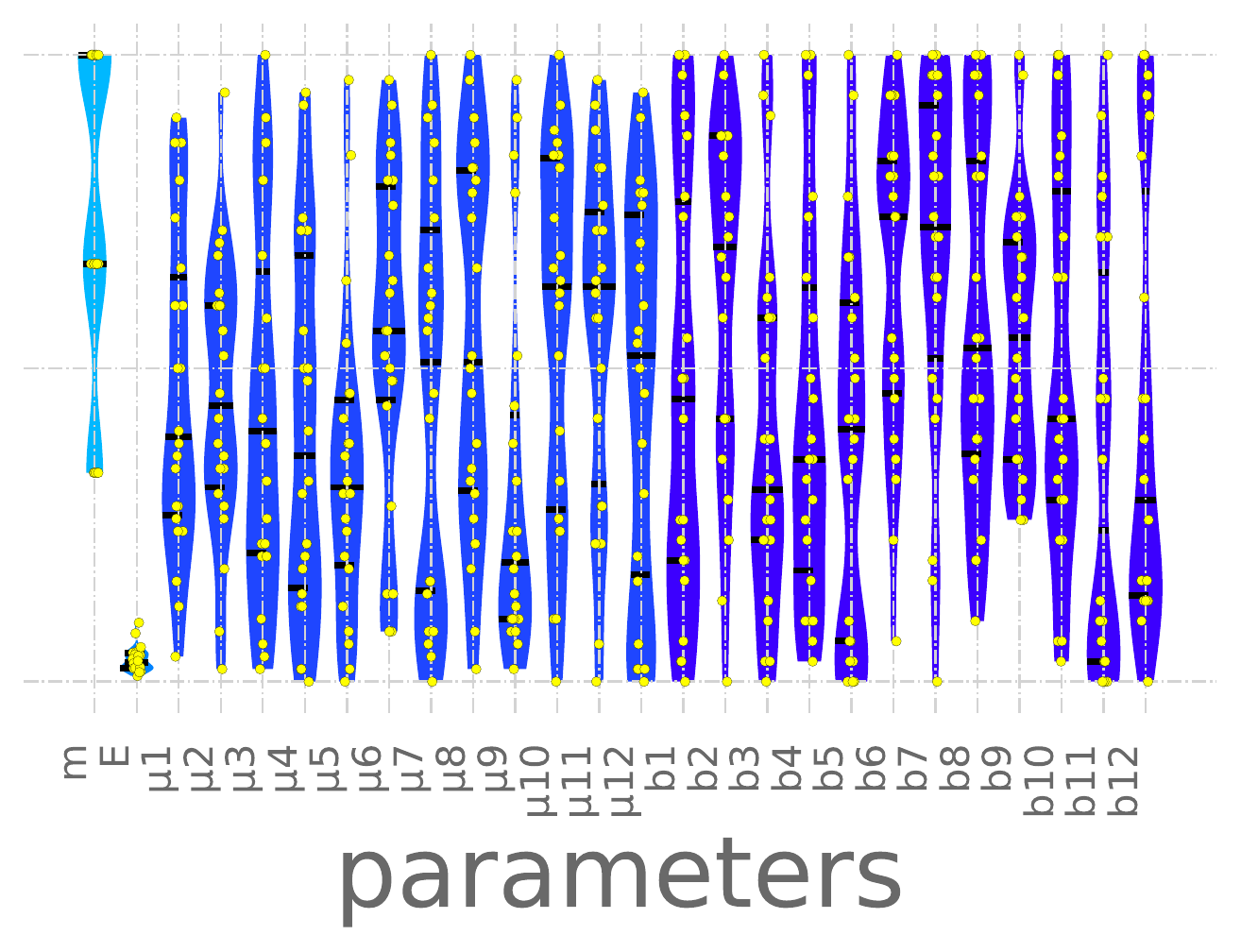}
         \caption{}
         \label{fig:params_2_dense}
     \end{subfigure}
     \hfill
     \begin{subfigure}[b]{0.32\textwidth}
         \centering
         \includegraphics[width=\textwidth]{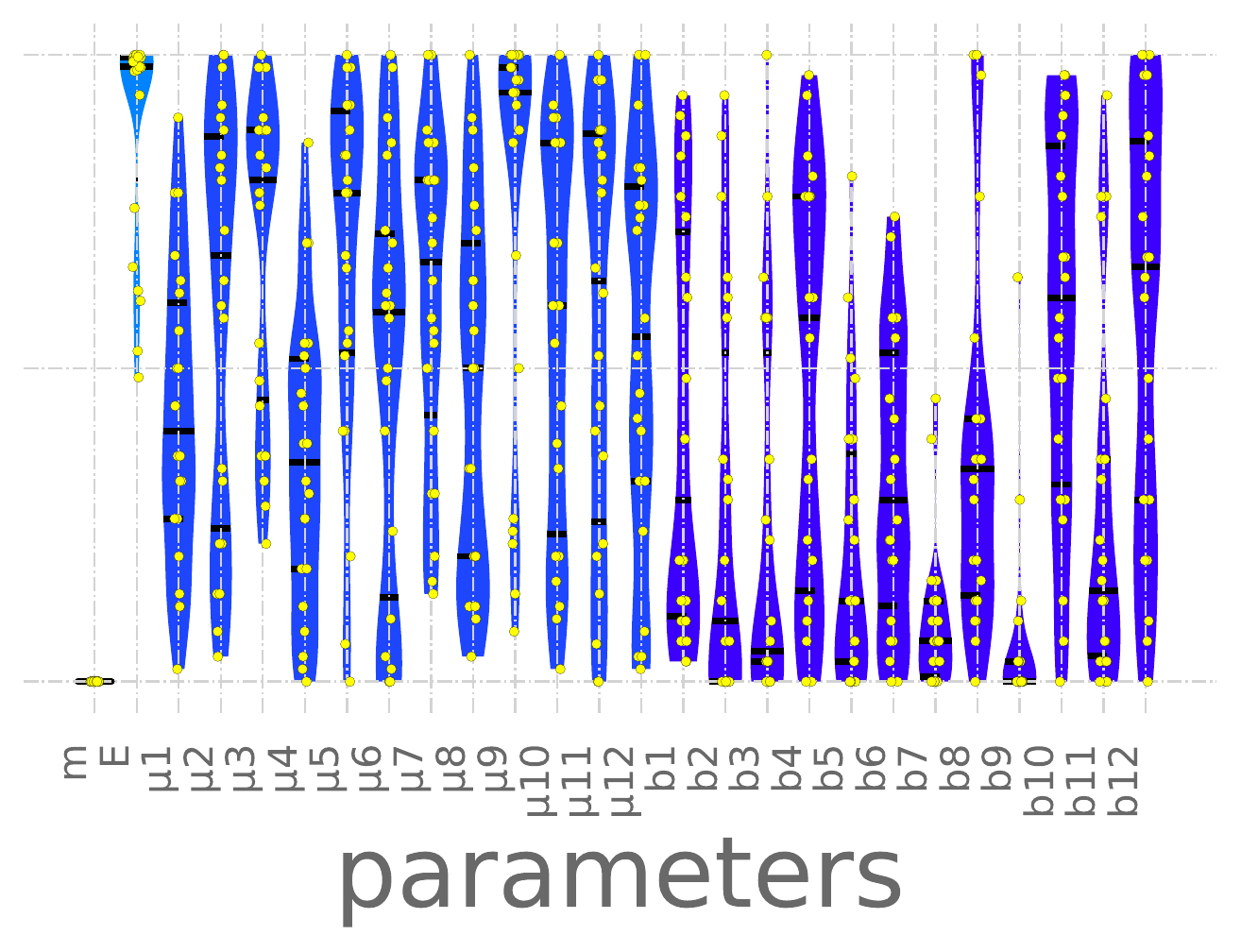}
         \caption{}
         \label{fig:params_3_dense}
     \end{subfigure}
     \hfill
     
     \begin{subfigure}[b]{0.32\textwidth}
         \centering
         \includegraphics[width=\textwidth]{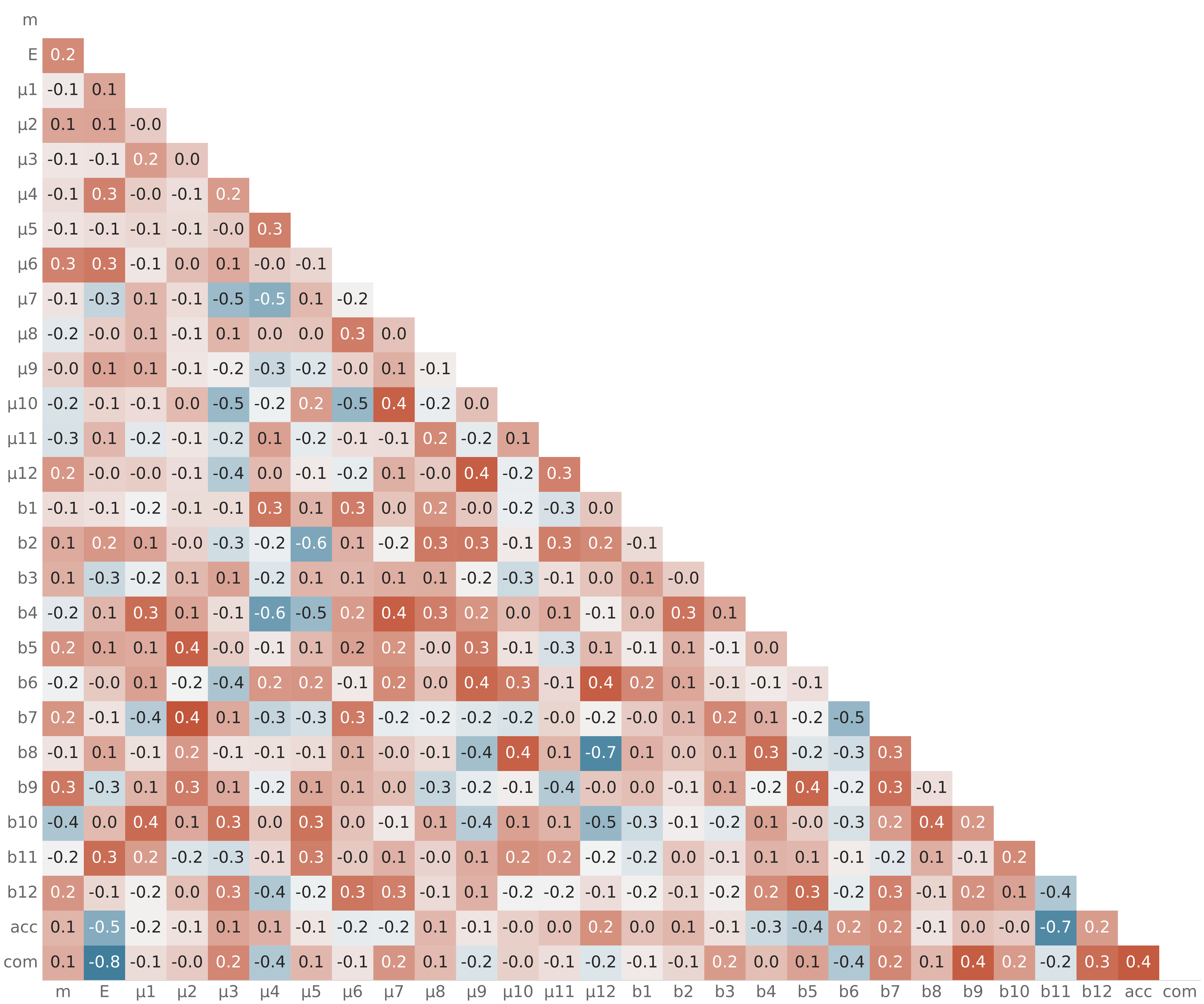}
         \caption{}
         \label{fig:heat_1_dense}
     \end{subfigure}
     \hfill
     \begin{subfigure}[b]{0.32\textwidth}
         \centering
         \includegraphics[width=\textwidth]{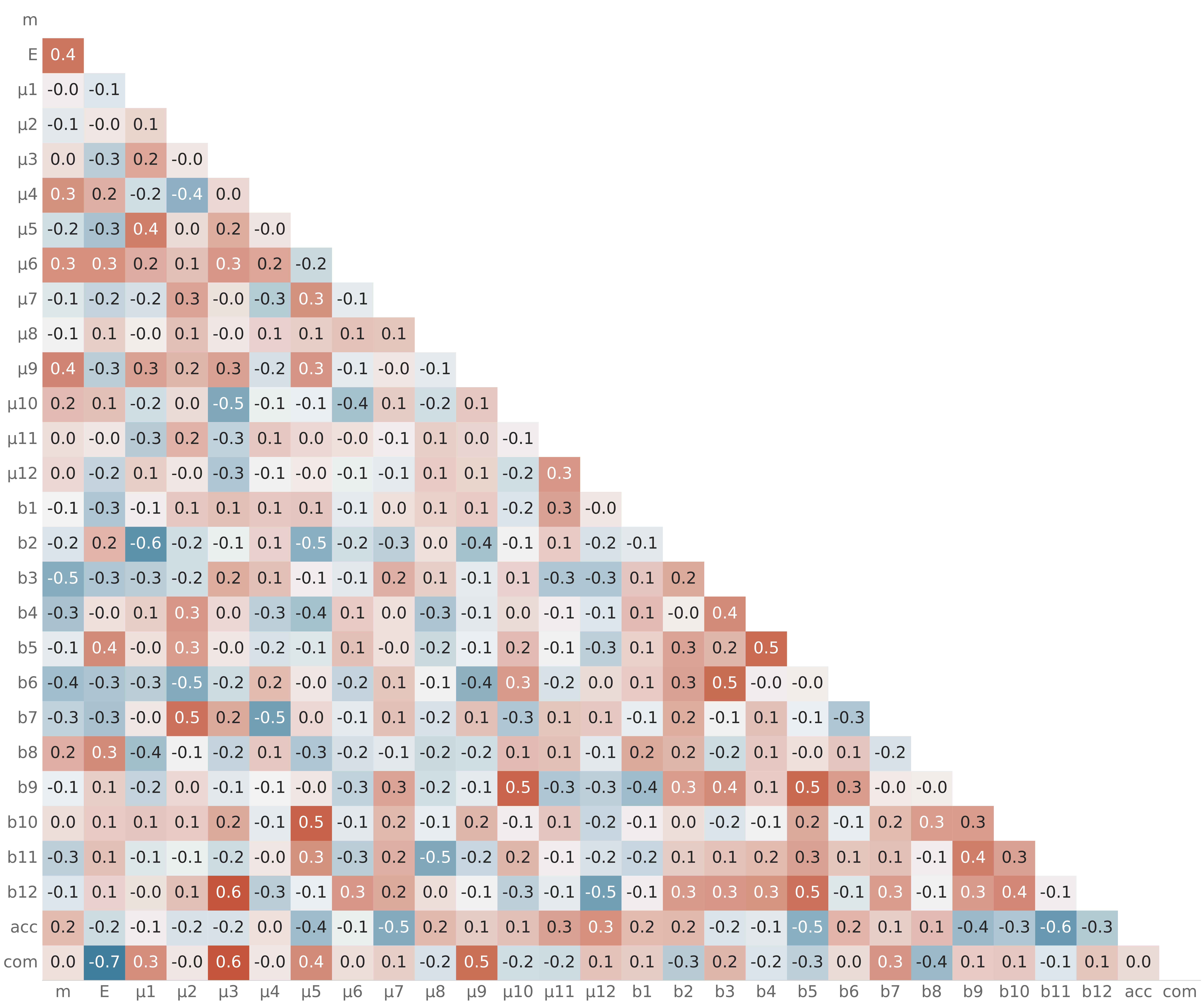}
         \caption{}
         \label{fig:heat_2_dense}
     \end{subfigure}
     \hfill
     \begin{subfigure}[b]{0.32\textwidth}
         \centering
         \includegraphics[width=\textwidth]{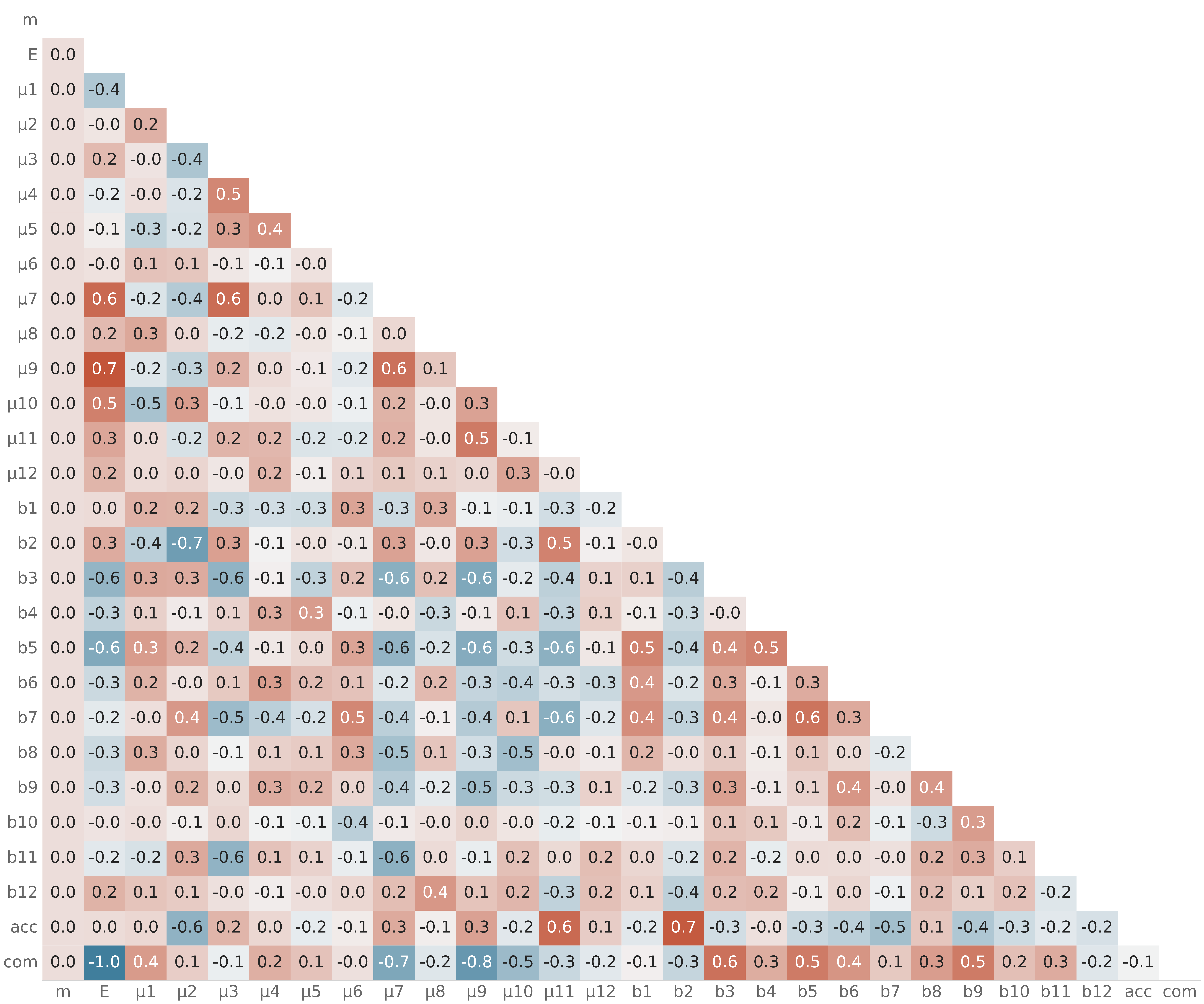}
         \caption{}
         \label{fig:heat_dense}
     \end{subfigure}  
     \hfill

     \begin{subfigure}[b]{0.32\textwidth}
         \centering
         \includegraphics[width=0.8\textwidth,height=2.2cm]{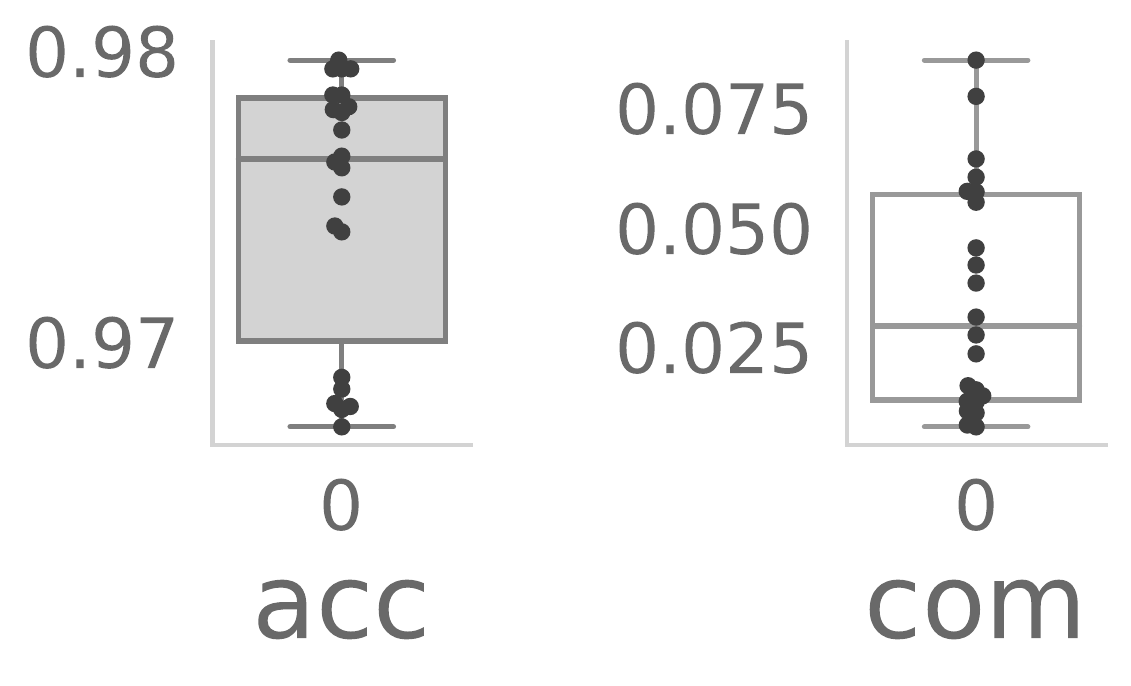}
         \caption{}
         \label{fig:heat_1_dense}
     \end{subfigure}
     \hfill
     \begin{subfigure}[b]{0.32\textwidth}
         \centering
         \includegraphics[width=0.8\textwidth,height=2.2cm]{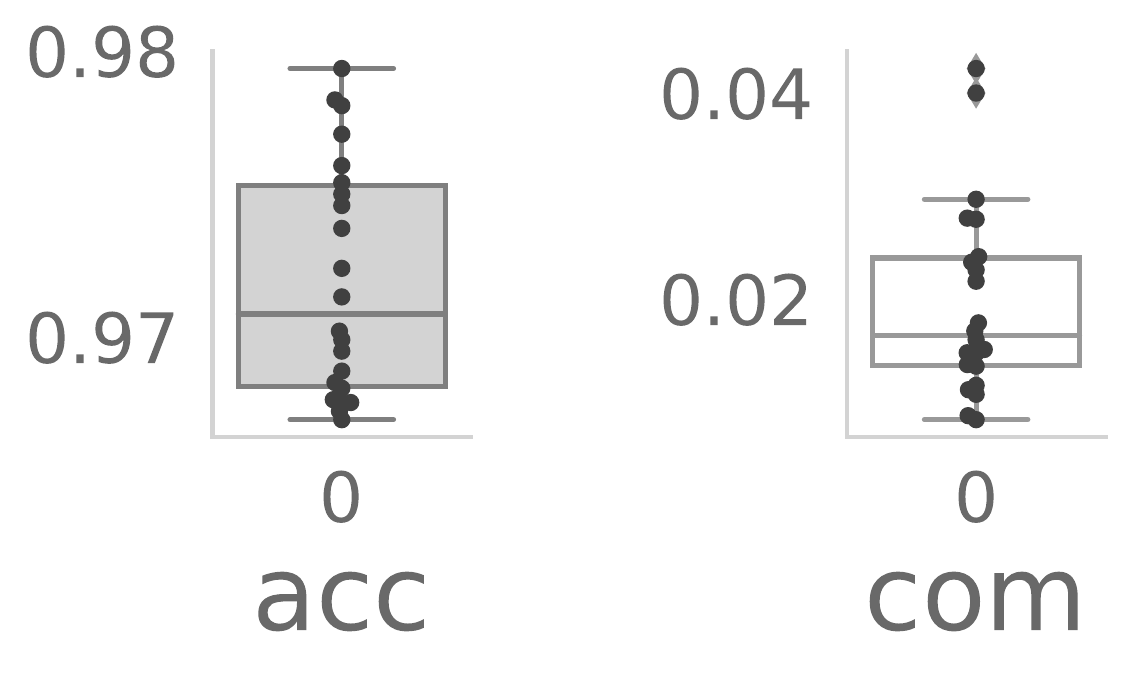}
         \caption{}
         \label{fig:heat_2_dense}
     \end{subfigure}
     \hfill
     \begin{subfigure}[b]{0.32\textwidth}
         \centering
         \includegraphics[width=0.8\textwidth,height=2.2cm]{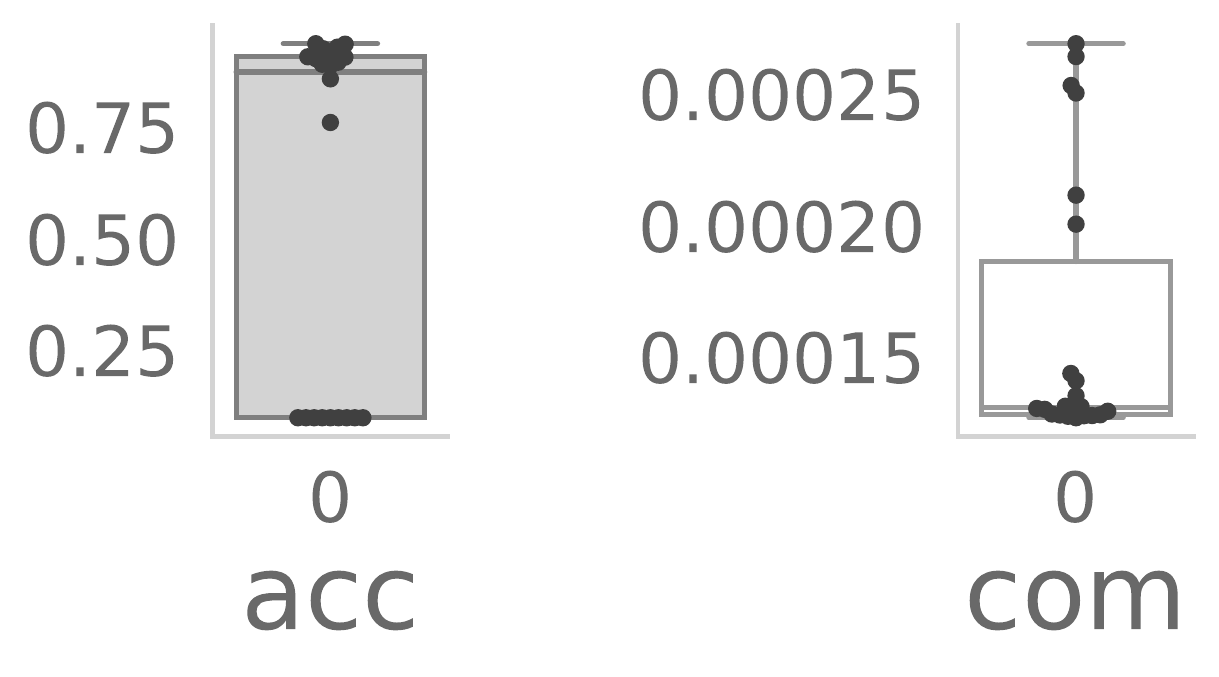}
         \caption{}
         \label{fig:heat_dense}
     \end{subfigure}  
     \vspace{-1em}
        \caption{Convolutional NN: (a)-(c) Normalised parameters of the individuals with higher accuracy in final Pareto fronts. (d)-(f) Normalised parameters of the individuals with best correlation accuracy vs. communications in final Pareto fronts. (g)-(i) Accuracy and communications mean, std, max, min and quartiles.}
         \label{fig:conv_stats}
\end{sidewaysfigure}

\vspace{-1.1em}

Fig. \ref{fig:dense_stats}(a) illustrates the median, first, and third quartiles of the distribution of the solution's parameters distribution for the individuals obtaining the best accuracy in each final Pareto of the 30 executions. Fig. \ref{fig:dense_stats}(b) and (c) do the same for the slope and lowest-communication solutions. Fig. \ref{fig:dense_stats}(a)-(c) concerns the fully-connected NN, while Fig. \ref{fig:conv_stats}(a)-(c) treats the convolutional NN. In Fig. \ref{fig:dense_stats}(a) and (b), the solutions with the best accuracy and slope solutions seem to share a similar distribution pattern of the parameters. This could be explained by the fact that both are generally located in the elbow of the Pareto, as it can be seen in Fig. \ref{Fig:dense2}. In both types of solutions a large part of the clients are frequently involved in the training, but to cope with this, a clear reduction of the communication rounds is noticed. Moreover, it seems that NSGA-II allows a homogeneous compression of weights in terms of number of bits used to code them as well as the percentage of weights being transferred. An explanation of such observation is an attempt of the algorithm of compensating the lack of communication (low $E$) by sending more precise ($b_i$) and complete weights ($\mu_i$) to help recover the model's accuracy. The results of the low-communication solution in Fig. \ref{fig:dense_stats}(c) are quite self-explanatory since the algorithm involve an almost null number of clients in the training, as well as a clear high rate of weight compression. This will probably induce a low-accuracy solution as it can be seen in Figure \ref{fig:dense_stats}(i). Moving now to the convolutional NN in Fig. \ref{fig:conv_stats}(a)-(c), NSGA-II produces solutions of different type where the number of participating clients is higher, while the weights' compression is non-homogeneous, but still meaningful. All the above stated-explanations can be mostly supported by Fig. \ref{fig:dense_stats}(c)-(i) and Fig. \ref{fig:conv_stats}(c)-(i). Another important conclusion that one can draw from Figs. \ref{Fig:dense1}-\ref{Fig:conv2}, is that the parameters that have more impact on the final communication are $m$ and $E$.  

\vspace{-1em}

%%%%%%%%%%%%%%%%%%%%%%%

\begin{figure}[h!]
     \centering
     \begin{subfigure}[b]{0.32\textwidth}
         \centering
         \includegraphics[width=\textwidth]{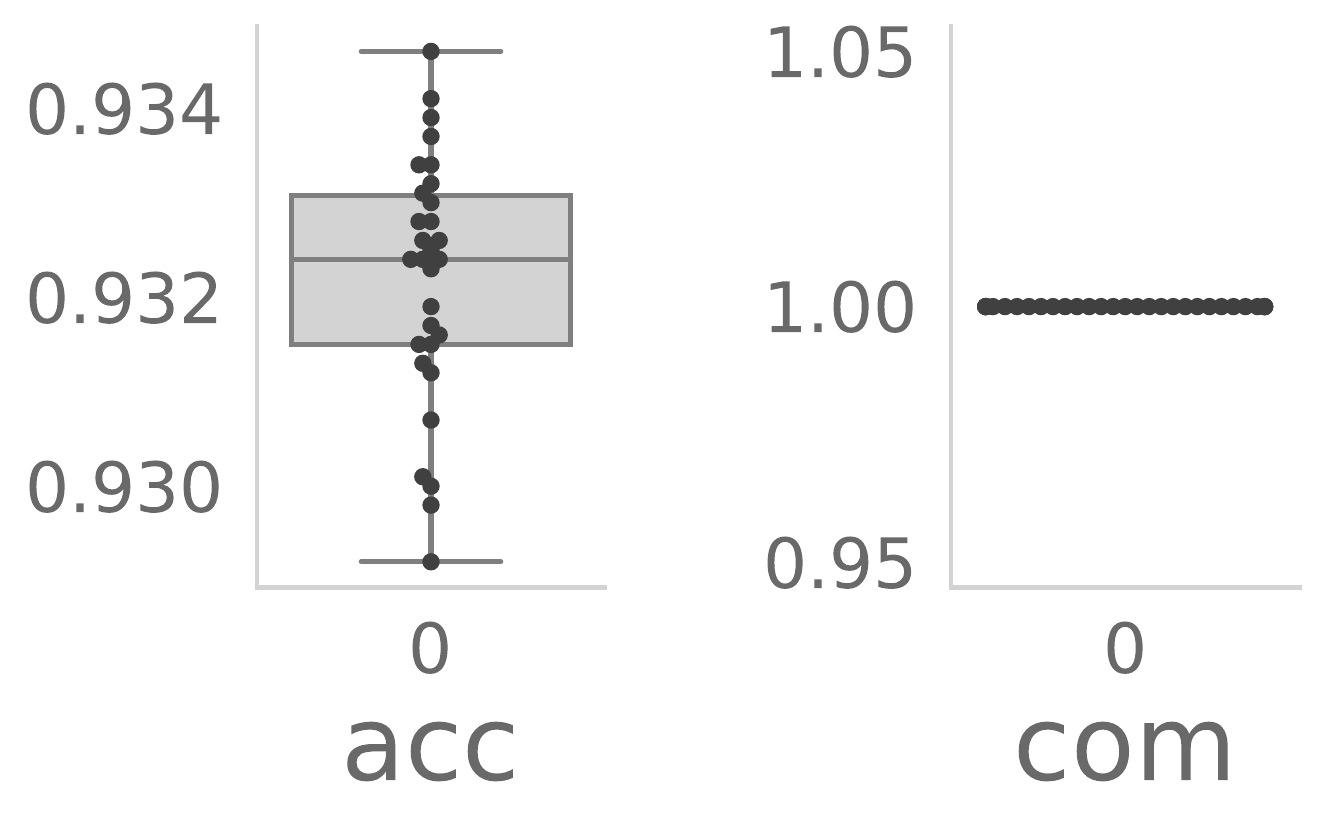}
         \caption{}
         \label{fig:params_1_full_dense}
     \end{subfigure}
     \hfill
     \begin{subfigure}[b]{0.32\textwidth}
         \centering
         \includegraphics[width=\textwidth]{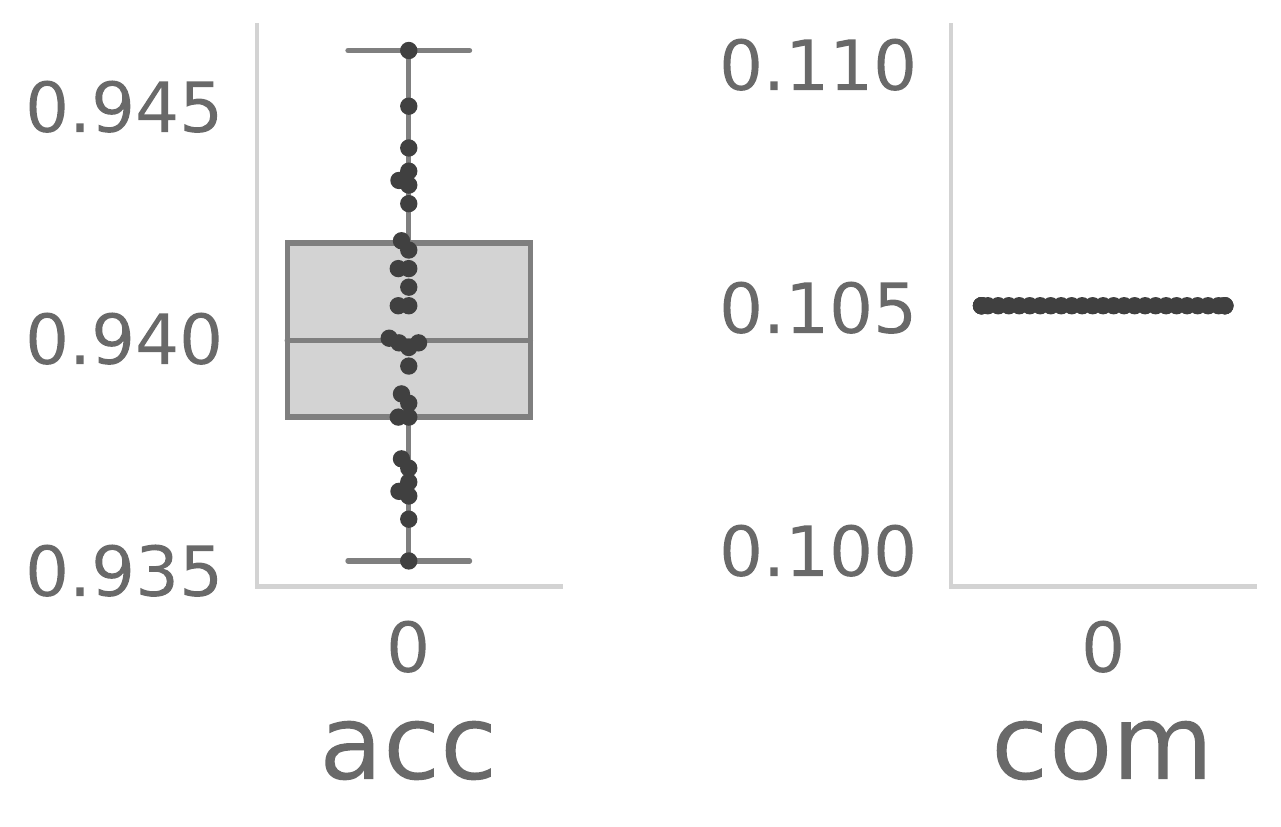}
         \caption{}
         \label{fig:params_2_full_dense}
     \end{subfigure}
     \hfill
     \begin{subfigure}[b]{0.32\textwidth}
         \centering
         \includegraphics[width=\textwidth]{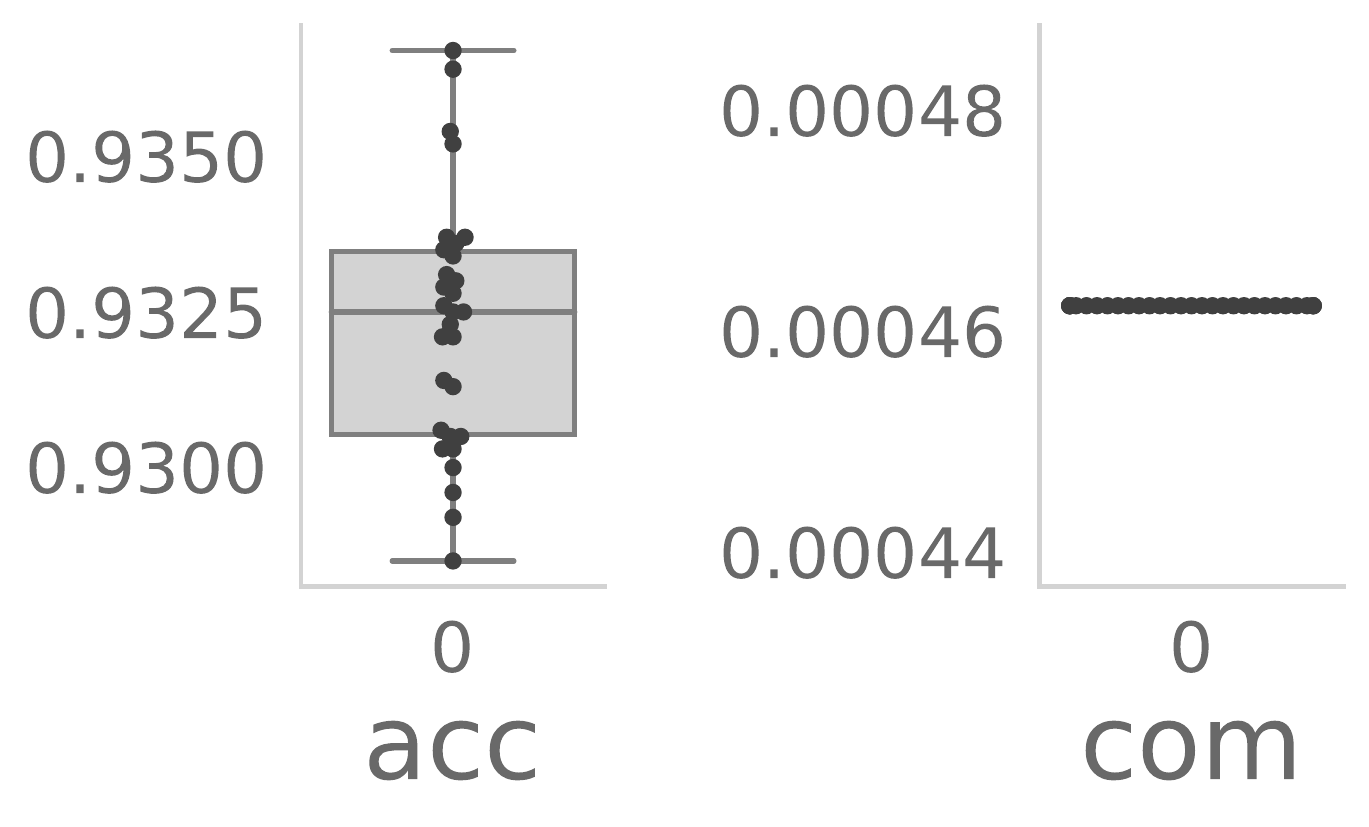}
         \caption{}
         \label{fig:params_3_full_dense}
     \end{subfigure}
     \hfill
     
     \begin{subfigure}[b]{0.32\textwidth}
         \centering
         \includegraphics[width=\textwidth]{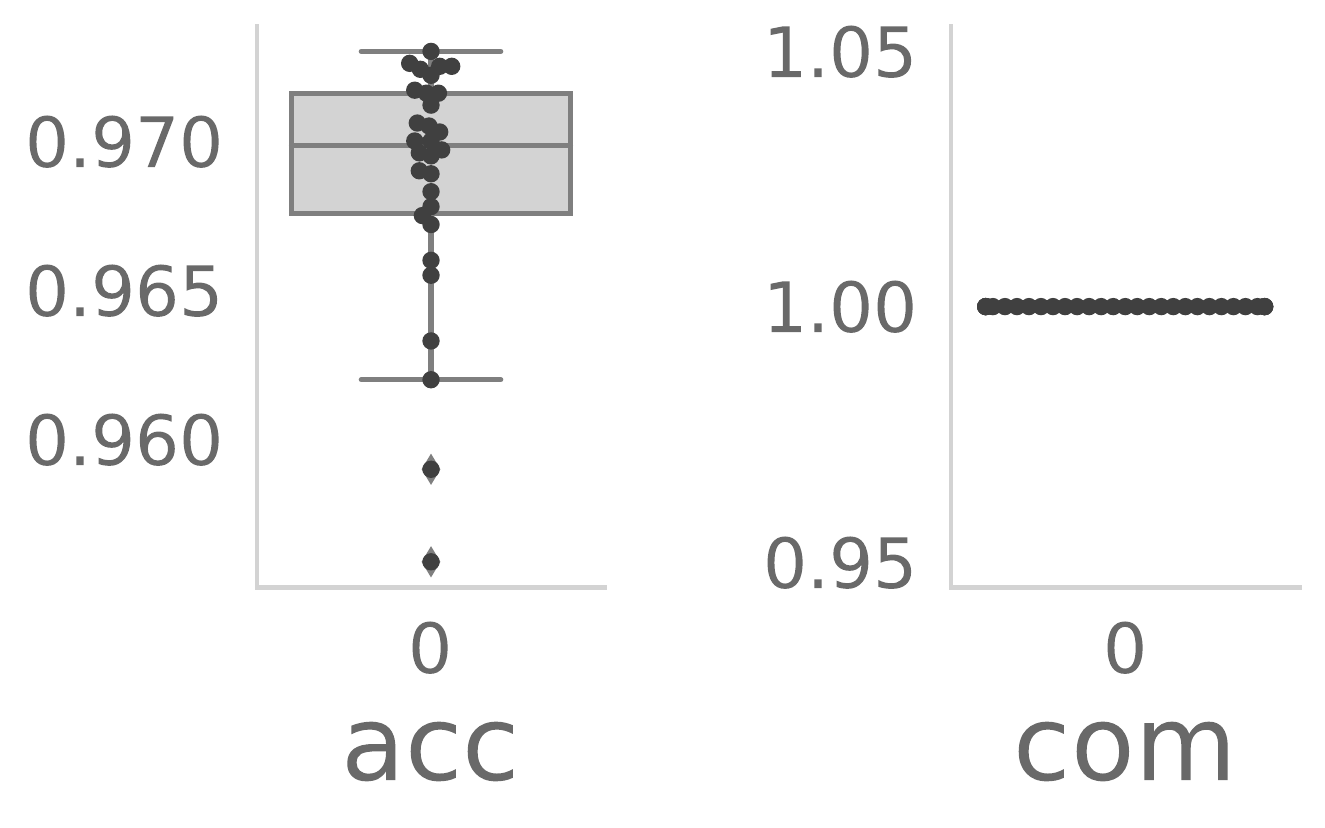}
         \caption{}
         \label{fig:params_1_full_conv}
     \end{subfigure}
     \hfill
     \begin{subfigure}[b]{0.32\textwidth}
         \centering
         \includegraphics[width=\textwidth]{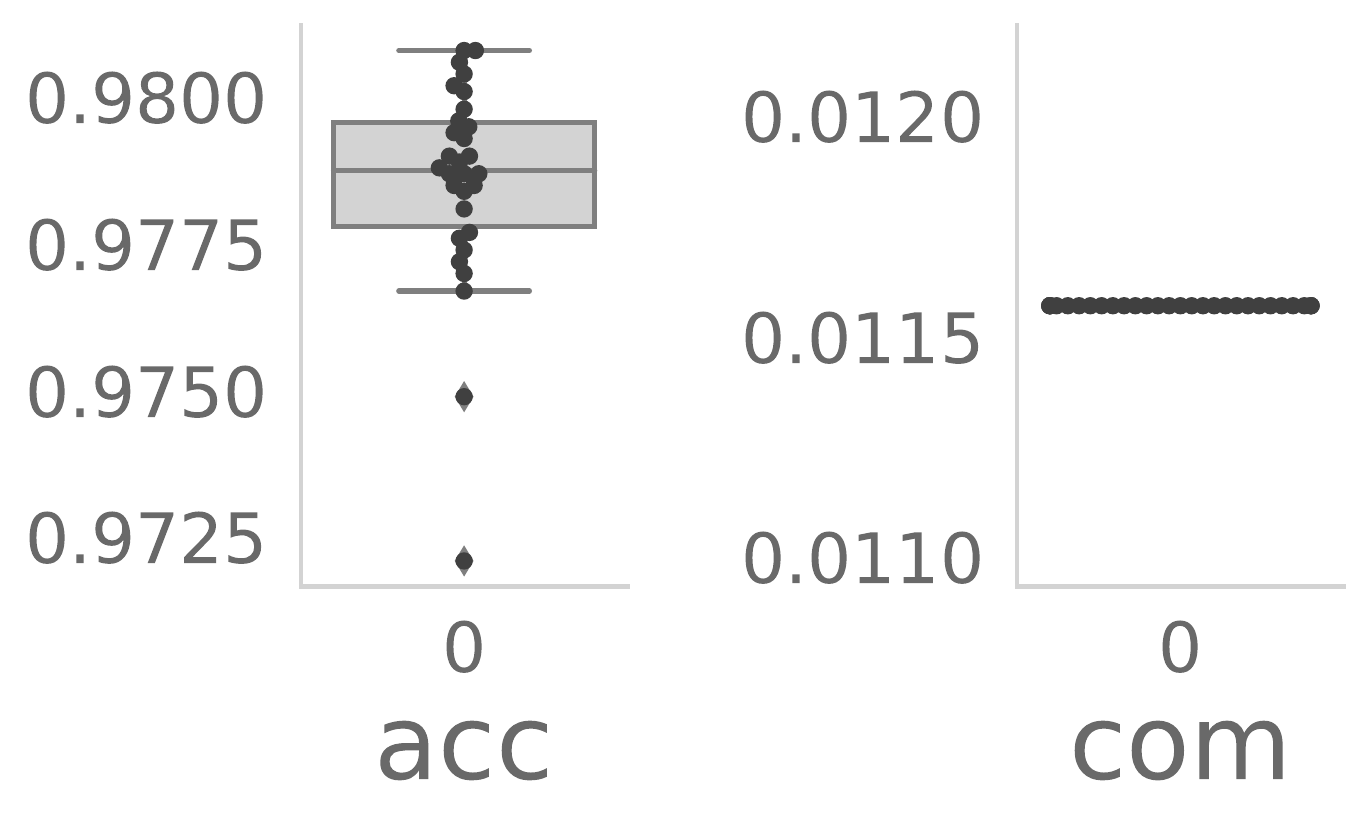}
         \caption{}
         \label{fig:params_2_full_conv}
     \end{subfigure}
     \hfill
     \begin{subfigure}[b]{0.32\textwidth}
         \centering
         \includegraphics[width=\textwidth]{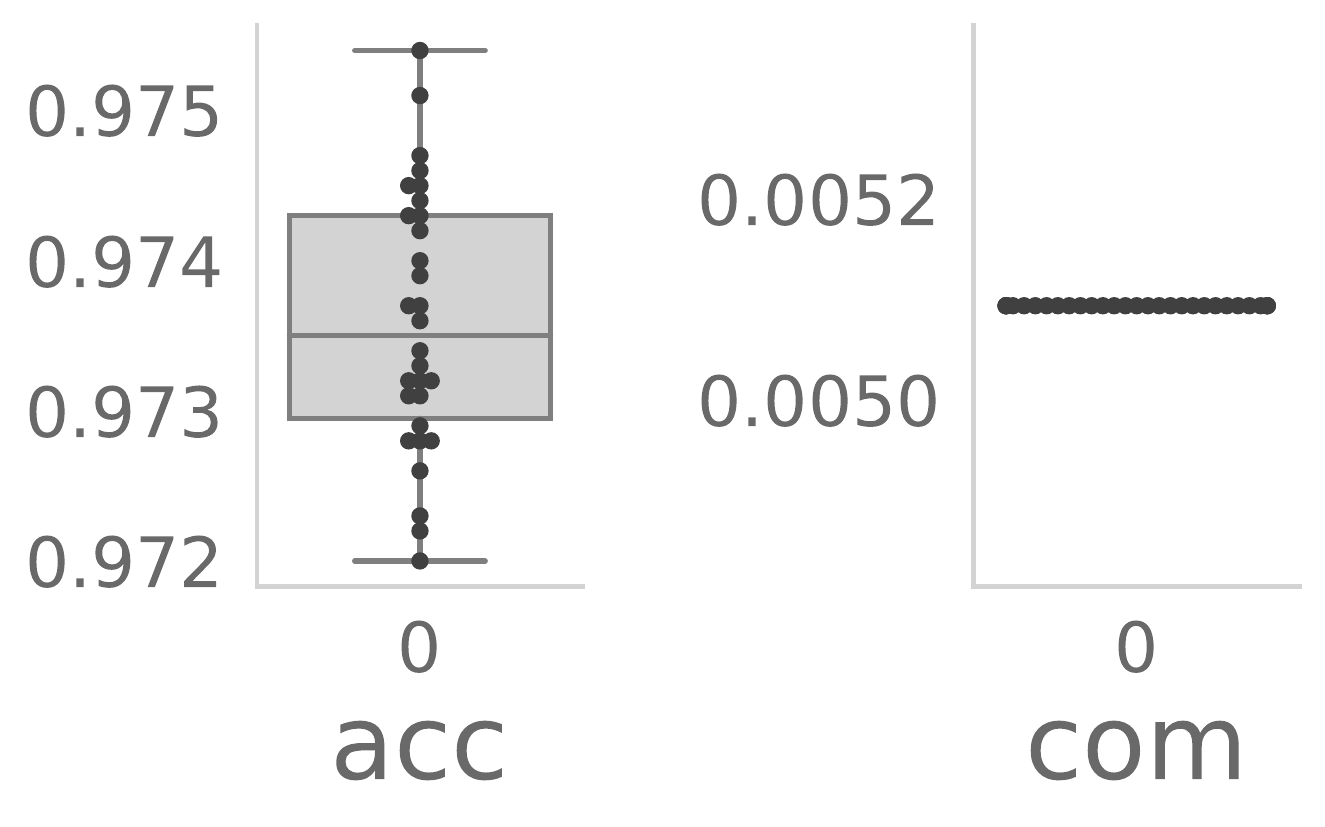}
         \caption{}
         \label{fig:params_3_full_conv}
     \end{subfigure}
     \hfill     

     \vspace{-1em}
        \caption{Comparing the solution obtained using a brute-force (100\%) communication with two solutions of the Pareto front obtained by the NSGA-II: (a)-(c) Fully-connected NN and (d)-(f) convolutional NN. }
         \label{fig:compar}
\end{figure}

\vspace{-2em}

Fig. \ref{fig:compar}(a)-(f) represents the statistical distribution of the communication load and  the model's accuracy obtained when using a brute-force (non-optimised) FedAvg that induces a 100\% of the communication load (Fig. \ref{fig:compar}(a) and (d)), with two solutions obtained by our proposal (i.e. NSGA-II), where one obtains the best accuracy (Fig. \ref{fig:compar}(b) and (e)) and one obtaining the lowest communication load (Fig. \ref{fig:compar}(c) and (f)). The results are obtained by executing the brute-force approach and the NSGA-II during 30 executions. The results are split into two categories: Fig. \ref{fig:compar}(a)-(c) for the fully-connected NN while, Fig. \ref{fig:compar}(d)-(f) for the convolutional one. One can note that for the case of fully-connected NN, our approach could reduce the communication overload by nearly 99\% while achieving the same accuracy as the one obtained by the brute-force-100\% communications. Taking the convolutional NN, the communication load has also been reduced by 99\% compared to the brute-force-100\% communication technique. Although the accuracy is not the same, we believe it is still acceptable.

\vspace{-1.4em}
\section{Conclusions and Perspectives \label{sec:cp}}
\vspace{-1em}
In this work, the FL-COP has been formulated as a multi-objective problem and solved using NSGA-II. As far as the authors' knowledge, this work is the first to \texttt{(I)} investigate the add-in that evolutionary computation can bring when solving this problem and, to do so, \texttt{(II)} considers both the model's and the properties of used devices. The experiments have been made by simulating a server/client architecture using 4 devices. Both convolutional and fully-connected neural networks of 12 and 3 layers with 887,530 and 33,400 weights, respectively, have been researched. The validation has been done on the \texttt{MNIST} dataset containing 70,000 images. The Experiments have shown that our approach could outperform the \texttt{FedAvg} algorithm using 100\% of communications. We could reduce the communication by 99\%, while maintaining an accuracy equal to the one obtained when using 100\% of communications (i.e. brute-force). As for future work, we aim at testing our proposal using physically distributed devices and larger benchmarks.
%\iffalse
\vspace{-2.3em}
\section*{Acknowledgments}
\vspace{-0.8em}
This research is partially funded by the Universidad de M\'alaga, Consejer\'ia de Econom\'ia y Conocimiento de la Junta de Andaluc\'ia and FEDER under grant number UMA18-FEDERJA-003 (PRECOG); under grant PID 2020-116727RB-I00 (HUmove) funded by MCIN/AEI/ 10.13039/501100011033; and TAILOR ICT-48 Network (No 952215) funded by EU Horizon 2020 research and innovation programme. José Ángel Morell is supported by an FPU grant from the Ministerio de Educación, Cultura y Deporte, Gobierno de España (FPU16/02595). The authors thank the Supercomputing and Bioinnovation Center (SCBI) for their provision of computational resources and technical support. The views expressed are purely those of the writer and may not in any circumstances be regarded as stating an official position of the European Commission. 
%\fi
\vspace{-1em}
\bibliography{bib}{}

\begin{thebibliography}{10}
\providecommand{\url}[1]{\texttt{#1}}
\providecommand{\urlprefix}{URL }
\providecommand{\doi}[1]{https://doi.org/#1}

\bibitem{entry_quantisation}
Alistarh, D., Grubic, D., Li, J.Z., Tomioka, R., Vojnovic, M.: {QSGD}:
  Communication-efficient {SGD} via gradient quantization and encoding. In:
  Proceedings of the 31st International Conference on Neural Information
  Processing Systems. p. 1707–1718. NIPS'17 (2017)

\bibitem{DPA02}
Deb, K., Pratap, A., Agarwal, S., Meyarivan, T.: A fast and elitist
  multiobjective genetic algorithm: {NSGA-II}. IEEE Trans. on Ev. Comp.
  \textbf{6}(2),  182--197 (2002)

\bibitem{lecun2015deep}
LeCun, Y., Bengio, Y., Hinton, G.: Deep learning. nature  \textbf{521},
  436--444 (2015)

\bibitem{mayer2020scalable}
Mayer, R., Jacobsen, H.A.: Scalable deep learning on distributed
  infrastructures: Challenges, techniques, and tools. ACM Computing Surveys
  \textbf{53}(1),  1--37 (2020)

\bibitem{mcmahan2017communication}
McMahan, B., Moore, E., Ramage, D., Hampson, S., y~Arcas, B.A.:
  Communication-efficient learning of deep networks from decentralized data.
  In: Artificial Intelligence and Statistics. pp. 1273--1282. PMLR (2017)

\bibitem{sheller2020federated}
Sheller, M.J., Edwards, B., Reina, G.A., Martin, J., Pati, S., Kotrotsou, A.,
  Milchenko, M., Xu, W., Marcus, D., Colen, R.R., et~al.: Federated learning in
  medicine: facilitating multi-institutional collaborations without sharing
  patient data. Scientific reports  \textbf{10}(1),  1--12 (2020)

\bibitem{tak2020federated}
Tak, A., Cherkaoui, S.: Federated edge learning: Design issues and challenges.
  IEEE Network  (2020)

\bibitem{entry_sparsification}
Wangni, J., Wang, J., Liu, J., Zhang, T.: Gradient sparsification for
  communication-efficient distributed optimization. In: Proceedings of
  32$^{nd}$ International Conference on Neural Information Processing Systems.
  p. 1306–1316 (2018)

\bibitem{xu2020ternary}
Xu, J., Du, W., Jin, Y., He, W., Cheng, R.: Ternary compression for
  communication-efficient federated learning. IEEE Transactions on Neural
  Networks and Learning Systems  (2020)

\bibitem{zhou2021communication}
Zhou, Y., Ye, Q., Lv, J.C.: Communication-efficient federated learning with
  compensated overlap-fedavg. IEEE Transactions on Parallel and Distributed
  Systems  (2021)

\end{thebibliography}

\bibliographystyle{splncs04}

\end{document}